%% file: egpaper_for_review.tex
\documentclass[10pt,twocolumn,letterpaper]{article}

\usepackage{cvpr}
\usepackage{times}
\usepackage{epsfig}
\usepackage{graphicx}
\usepackage{subfigure} 
\usepackage{amssymb}
\usepackage[utf8]{inputenc} 
\usepackage[T1]{fontenc}    
\usepackage{url}            
\usepackage{booktabs}       
\usepackage{amsfonts}       
\usepackage{nicefrac}       
\usepackage{microtype}      
\usepackage{amsmath,amsthm}
\usepackage{algorithm}
\usepackage{algorithmic}
\usepackage{cite}
\usepackage{verbatim}
\usepackage{pdfpages}
\newtheorem{definition}{Definition}
\newtheorem{reduction}{Reduction}
\newtheorem{theorem}{Theorem}
\newtheorem{corollary}{Corollary}[section]

\usepackage{lipsum}

\newcommand\blfootnote[1]{%
  \begingroup
  \renewcommand\thefootnote{}\footnote{#1}%
  \addtocounter{footnote}{-1}%
  \endgroup
}

\usepackage[pagebackref=true,breaklinks=true,letterpaper=true,colorlinks,bookmarks=false]{hyperref}

\cvprfinalcopy 

\def\cvprPaperID{7550} 

\ifcvprfinal\fi
\begin{document}

\title{Learning to Select Base Classes for Few-shot Classification}

\author{Linjun Zhou$^{1, 2}$ \qquad Peng Cui$^{1, *}$ \qquad Xu Jia$^{2, *}$ \qquad Shiqiang Yang$^1$ \qquad Qi Tian$^2$\\
$^1$Tsinghua University \qquad $^2$Noah's Ark Lab, Huawei Technologies\\
\tt\small zhoulj16@mails.tsinghua.edu.cn,\quad \tt\small cuip@tsinghua.edu.cn\\ \tt\small jiayushenyang@gmail.com, \quad \tt\small yangshq@tsinghua.edu.cn, \quad \tt\small tian.qi1@huawei.com}

\maketitle
\thispagestyle{empty}

\begin{abstract}
   \blfootnote{* Co-corresponding authors.}
   Few-shot learning has attracted intensive research attention in recent years. Many methods have been proposed to generalize a model learned from provided base classes to novel classes, but no previous work studies how to select base classes, or even whether different base classes will result in different generalization performance of the learned model. In this paper, we utilize a simple yet effective measure, the Similarity Ratio, as an indicator for the generalization performance of a few-shot model. We then formulate the base class selection problem as a submodular optimization problem over Similarity Ratio. We further provide theoretical analysis on the optimization lower bound of different optimization methods, which could be used to identify the most appropriate algorithm for different experimental settings. The extensive experiments on ImageNet \cite{imagenet_cvpr09}, Caltech256 \cite{griffin2007caltech} and CUB-200-2011 \cite{WahCUB_200_2011} demonstrate that our proposed method is effective in selecting a better base dataset.
\end{abstract}

\input{article.tex}

{\small
\bibliographystyle{ieee_fullname}
\bibliography{egbib}
}

\appendix
\cleardoublepage


\section*{Supplementary material}

# Appendix of Learning to Select Base Classes for Few-shot Classification

## 1. Proof of the Main Theories

### 1.1. Proof for Corollary 4.1

**Lemma 1.** $\forall n \in N, g_n : 2^{B_u} \to \mathbb{R}_{\geq 0}, g_n(U) := M^K(f(c_n, \{c_{B_s}, c_U\}))$ *is a submodular function.*

*Proof.* $\forall A \subseteq B \subseteq B_U$, let $u \in B_U \backslash B$, define the top-K similar classes with class n in $B_s \cup A$ and $B_s \cup B$ are $K_A$ and $K_B$ separately, we also define that after adding class u to both $A$ and $B$, the top-K similar classes become $K_A^u$ and $K_B^u$. Next, we discuss four cases:

(1) $u \in K_A^u$ but $u \in K_B^u$: In this case, $g_n(A + u) - g_n(A) = f(c_n, c_u) - \min_{x \in K_A} f(c_n, c_x)$ and $g_n(B + u) - g_n(B) = f(c_n, c_u) - \min_{x \in K_B} f(c_n, c_x)$. As $(B_s \cup A) \subseteq (B_s \cup B)$, there must be $\min_{x \in K_A} f(c_n, c_x) \leq \min_{x \in K_B} f(c_n, c_x)$. Thus we have $g_n(A + u) - g_n(A) \geq g_n(B + u) - g_n(B)$.

(2) $u \in K_A^u$ but $u \notin K_B^u$: In this case, easy to show that $g_n(A + u) - g_n(A) > 0 = g_n(B + u) - g_n(B)$.

(3) $u \notin K_A^u$ but $u \in K_B^u$: This case will not exist, as it represents that $f(c_n, c_u) \leq \min_{x \in K_A} f(c_n, c_x)$ and $f(c_n, c_u) \geq \min_{x \in K_B} f(c_n, c_x)$. This will induce a contradictory to $\min_{x \in K_A} f(c_n, c_x) \leq \min_{x \in K_B} f(c_n, c_x)$.

(4) $u \notin K_A^u$ but $u \notin K_B^u$: In this case, easy to show that $g_n(A + u) - g_n(A) = g_n(B + u) - g_n(B) = 0$.

In conclusion, $\forall A \subseteq B \subseteq B_U, u \in B_U \backslash B$, we have $g_n(A + u) - g_n(A) \geq g_n(B + u) - g_n(B)$, which demonstrates that $g_n(\cdot)$ is a submodular function. □

**Corollary 1.** *(Corollary 4.1 in original paper) Considering optimization problem 3 (in original paper), when $\lambda = 0$, Problem 3 is equivalent to a submodular monotone non-decreasing optimization with exact cardinality constraint and when $\lambda > 0$, Problem 3 is equivalent to a submodular optimization with exact cardinality.*

*Proof.* When $\lambda = 0$, by Lemma 1 and the property of the additivity of submodular function that if $f$ and $g$ are both submodular, then $h = f + g$ is also submodular, easy to show that the optimization function is submodular. Easy to show that $g_n(U)$ is also monotone non-decreasing, so Problem 3 with $\lambda = 0$ is a submodular monotone non-decreasing optimization with exact cardinality constraint. Also, the regularizer term of the optimization function

$$R(U) = \sum_{n \in N} \frac{1}{|B_s| + m} \sum_{u \in B_s \cup U} f(c_n, c_u)$$

is a modular function satisfying $R(A + u) - R(A) = R(B + u) - R(B), \forall A \subseteq B \subseteq V$. By the property of submodular function, the whole optimization function 3 with $\lambda > 0$ is also a submodular function (but not monotone non-decreasing). □

### 1.2. Proof for Theorem 1

**Theorem 1.** *For $B_s = \emptyset$ and $\lambda = 0$, when $m \geq K \cdot |N|$, using Greedy on Novel Class to solve for optimization problem 3, the solution will be optimal.*

*Proof.* The maximum number of base classes for top-K most similar classes with each novel class is $K \cdot |N|$, thus when $m \geq K \cdot |N|$, a greedy algorithm on finding top-K most similar classes for each novel class is optimal. □

### 1.3. Proof for Theorem 2

**Theorem 2.** *For $B_s = \emptyset$ and $\lambda = 0$, using Greedy on Target Function to solve for optimization problem 3, let $h(\cdot)$ be the optimization function, and let Q be*

$$Q = \mathbb{E}_{u \sim Uniform(B), v \sim Uniform(N)}(f(c_u, c_v))$$

*representing for the average similarity between base classes and novel classes, we have $h(U) \geq (1 - 1/e) \cdot h(OPT) + 1/e \cdot Q$.*

*Proof.* Let us suppose $A_i$ denotes the chosen subset after greedy step $i$. Let function $\gamma(u) = \frac{1}{N} \sum_{n \in N} f(c_n, c_u)$. According to the greedy algorithm, $A_K$ should be top-k elements in $B_u$ maximizing $\gamma(u)$. Easy to show that $h(A_K) = \frac{1}{K} \sum_{u \in A_K} \gamma(u) \geq Q$.

[3] shows that for submodular monotone non-decreasing problem, we have

$$h(OPT) - h(A_i) \leq (1 - 1/k) \cdot (h(OPT) - h(A_{i-1})),$$

Combining the inequality for every $K \leq i \leq m$ and take limitations we have

$$h(U) = h(A_m) \geq (1 - 1/e) \cdot h(OPT) + 1/e \cdot h(A_K) \geq (1 - 1/e) \cdot h(OPT) + 1/e \cdot Q.$$

$\square$

### 1.4. Proof for Theorem 3

**Theorem 3.** *Let $S \subseteq B_u$ is a random set, with each element $v$ in $B_u$ i.i.d sampled with probability $(x \wedge (B_u - u))_v$. For each novel class $n \in N$, we sort the similarity function $f(c_n, c_b)$ for every base class $b \in B$ in descent order, denoting as $q_{n,[1]}, q_{n,[2]}, \cdots q_{n,[|B|]}$. Similarly, we also sort the similarity function for every base class in $S \cup B_s$ in descent order, denoting as $s_{n,[1]}, s_{n,[2]}, \cdots s_{n,[|S|+|B_s|]}$, then we have:*

$$F(x \vee u) - F(x \wedge (B_u - u))$$
$$= \frac{1}{|N| \cdot K} \sum_{n \in N} \sum_{i=1}^{|B|} P(s_{n,[K]} = q_{n,[i]}) \max(f(c_n, c_u) - q_{n,[i]}, 0) \quad (1)$$
$$- \lambda \cdot \frac{1}{|N| \cdot m} \sum_{n \in N} f(c_n, c_u)$$

*Proof.*

$$F(x \vee u) - F(x \wedge (B_u - u))$$
$$= \sum_{\substack{S \subseteq B_u \setminus u}} h(S + u) \prod_{\substack{v \in S \\ v \neq u}} x_v \prod_{\substack{v \notin S \\ v \neq u}} (1 - x_v) \cdot 1 - \sum_{\substack{S \subseteq B_u \setminus u}} h(S) \prod_{\substack{v \in S \\ v \neq u}} x_v \prod_{\substack{v \notin S \\ v \neq u}} (1 - x_v) \cdot 1$$
$$= \sum_{S \subseteq B_u \setminus u} (h(S + u) - h(S)) \prod_{\substack{v \in S \\ v \neq u}} x_v \prod_{\substack{v \notin S \\ v \neq u}} (1 - x_v)$$
$$= \sum_{S \subseteq B_u \setminus u} (\frac{1}{|N| \cdot K} \sum_{n \in N} \max(f(c_n, c_u) - s_{n,[K]}, 0)) \prod_{\substack{v \in S \\ v \neq u}} x_v \prod_{\substack{v \notin S \\ v \neq u}} (1 - x_v)$$
$$- \lambda \cdot \frac{1}{|N| \cdot m} \sum_{n \in N} f(c_n, c_u) \sum_{S \subseteq B_u \setminus u} (\prod_{\substack{v \in S \\ v \neq u}} x_v \prod_{\substack{v \notin S \\ v \neq u}} (1 - x_v))$$
$$= \frac{1}{|N| \cdot K} \sum_{n \in N} \sum_{i=1}^{|B|} P(s_{n,[K]} = q_{n,[i]}) \max(f(c_n, c_u) - q_{n,[i]}, 0)$$
$$- \lambda \cdot \frac{1}{|N| \cdot m} \sum_{n \in N} f(c_n, c_u)$$

$\square$

### 1.5. Proof for Equation 7 (in Original Paper)

The only unknown term $P(s_{n,[K]} = q_{n,[i]})$ for $n \in N$ could be solved using dynamic programming in $O(K \cdot |B| \cdot |N|)$ time complexity by the following two recursion equations:

$$\begin{cases} P(s_{n,[j]} \geq q_{n,[i]}) = (1 - x_{[i]}) \cdot P(s_{n,[j]} \geq q_{n,[i-1]}) + x_{[i]} \cdot P(s_{n,[j-1]} \geq q_{n,[i-1]}) \ for \ [i] \in B_u \\ P(s_{n,[j]} \geq q_{n,[i]}) = P(s_{n,[j-1]} \geq q_{n,[i-1]}) \ for \ [i] \in B_s \end{cases} \quad (2)$$

$$P(s_{n,[j]} = q_{n,[i]}) = P(s_{n,[j]} \geq q_{n,[i]}) - P(s_{n,[j]} \geq q_{n,[i-1]}) \tag{3}$$

*Proof.* Equation 3 is obvious. Below we give the proof of 2, for the case of $[i] \in B_u$:

$$P(s_{n,[j]} = q_{n,[i]})$$
$$= \{P(s_{n,[j-1]} \geq q_{n,[i-1]}) - \sum_{m=1}^{i-1} P(s_{n,[j]} = q_{n,[m]})\} \cdot x_{[i]}$$
$$= \{P(s_{n,[j-1]} \geq q_{n,[i-1]}) - \sum_{m=1}^{i-1} (P(s_{n,[j]} \geq q_{n,[m]}) - P(s_{n,[j]} \geq q_{n,[m-1]}))\} \cdot x_{[i]}$$
$$= \{P(s_{n,[j-1]} \geq q_{n,[i-1]}) - P(s_{n,[j]} \geq q_{n,[i-1]})\} \cdot x_{[i]}$$

Plug Equation 3 to the equation above and that will be Equation 2.

Note that the vector $x$ could also be seen as an extension form in $[0,1]^{|B|}$: $x_{[i]}$ represents for the probability of element $[i]$ being selected, when using these two equations, if $[i] \in B_s$ we set $x_{[i]} = 1$; if $[i] = v \in B_u \backslash u$, we set $x_{[i]} = x_v$; and if $[i] = u$ we set $x_{[i]} = 0$. Hence for $[i] \in B_s$ we have $x_{[i]} = 1$ and plug into the first equation of 2 to obtain the second equation. □

### 1.6. Proof for Theorem 4

**Theorem 4.** *For $B_s = \emptyset$ and $\lambda > 0$, using a combination of Random Greedy Algorithm and Continuous Double Greedy Algorithm to solve for optimization problem 3, let $h(\cdot)$ be the optimization function, and let $Q$ be*

$$Q = \mathbb{E}_{u \sim Uniform(B), v \sim Uniform(N)}(f(c_u, c_v))$$

*representing for the average similarity between base classes and novel classes, and let $r$ be the cardinality of $B_u$, we have*

$$\mathbb{E}(h(U)) \geq max(\frac{1 - m/er}{e} \cdot h(OPT) + C_1 \cdot Q, (1 + \frac{r}{2\sqrt{(r-m)m}})^{-1} \cdot h(OPT) + C_2 \cdot Q)$$

*For $0 < \lambda < \frac{1}{e-1}$, we have $C_1 = \frac{1}{e} + (1 - \frac{1}{e})\frac{m}{r} - (1 - \frac{1}{e}) \cdot \lambda > 0$ and $C_2 = \frac{(1-\lambda)r}{2\sqrt{(r-m)m+r}} - \epsilon \geq \frac{1}{2}(1 - \lambda) > 0$.*

*Proof.* 1. For random greedy algorithm, our proof follows the Lemma 4.7 and Lemma 4.8 in [2] with slight differences. We suggest the readers read the proof of [2] foreahead. The first difference is that $h(B_u)$ may be negative, and it should not be taken away while calculating $\mathbb{E}(h(A_{i-1} \cup M'_i))$. Considering $h(B_u) < 0$ we have:

$$\frac{m - X_{i-1}}{r} \cdot h(B_u) \geq \frac{m}{r} \cdot h(B_u) = \frac{m}{r}(1 - \lambda \cdot \frac{r}{m}) \cdot Q \tag{4}$$

The inequality follows by the definition of $X_{i-1}$: $X_{i-1} = |OPT \backslash A_i| \geq 0$. And this term should be added to RHS of Lemma 4.7 in [2] with a $m^{-1}$ coefficient according to the process of proof. Thus Lemma 4.7 could be rewritten in our problem as: for every $K \leq i \leq m$:

$$\mathbb{E}(h_{u_i}(A_{i-1})) \geq \frac{[r/m - 1 + (1 - 1/m)^{i-1}] \cdot (1 - 1/k)^{i-1}}{n} \cdot h(OPT)$$
$$- \frac{\mathbb{E}(h(A_{i-1}))}{k} + \frac{1}{r}(1 - \lambda \cdot \frac{r}{m}) \cdot Q - E_{i-1}$$

The second difference is that we start our algorithm from $i = K$ and similar to Theorem 1 we have

$$h(A_K) = (\frac{1}{K} - \frac{\lambda}{m}) \sum_{u \in A_K} \gamma(u) \geq (1 - \frac{\lambda \cdot K}{m}) \cdot Q. \tag{5}$$

Thus compared to Lemma 4.8 in [2], we need to add two terms related to Equation 4 and 5. After repeated applications of Lemma 4.7, the term related to 4 is calculated by:

$$\lim_{\substack{K \to 0 \\ m \to +\infty}} \sum_{i=K}^{m} (1 - \frac{1}{m})^i \cdot (\frac{1}{r}(1 - \lambda \cdot \frac{r}{m}) \cdot Q) = (1 - \frac{1}{e})\frac{m}{r}(1 - \lambda\frac{r}{m}) \cdot Q.$$

And the term related to 5 is calculated by:

$$\lim_{\substack{K \to 0 \\ m \to +\infty}} (1 - \frac{1}{m})^{m-K} h(A_k) \geq \frac{1}{e} \cdot Q.$$

Thus combine these term with the coefficient $h(OPT)$ unchanged we could prove that:

$$C_1 = \frac{1}{e} + (1 - \frac{1}{e})\frac{m}{r} - (1 - \frac{1}{e}) \cdot \lambda$$

And for $\lambda > 1/(e-1)$, $C_1$ guarantees to be non-negative.

2. For double continuous greedy algorithm, refer to the Theorem 3.2 in [2] with some deformation we have:

$$h(U) \geq \frac{h(OPT) + \frac{1}{2}(\sqrt{\frac{r-m+K}{m-K}})h(A_K) + \sqrt{\frac{m-K}{r-m+K}}h(B_u))}{1 + \frac{1}{2}\frac{r}{\sqrt{(r-m+K)(m-K)}}} \quad (6)$$

From Theorem 1, we could conclude that $h(A_K) = (\frac{1}{K} - \frac{\lambda}{m})\sum_{u \in A_K} \gamma(u) \geq (1 - \frac{\lambda \cdot K}{m}) \cdot Q$. Also, easy to show that $h(B_u) \geq (1 - \frac{\lambda \cdot r}{m}) \cdot Q$. Thus we could put these two inequality to Equation 6 and as $K << m$ and $K << r$, we could omit the term with K. Then Equation 6 is equivalent to the inequality below:

$$h(U) \geq (1 + \frac{r}{2\sqrt{(r-m)m}})^{-1} \cdot h(OPT) + C_2 \cdot Q) \quad (7)$$

And we have:

$$C_2 = (1 + \frac{r}{2\sqrt{(r-m)m}})^{-1} \cdot \frac{1}{2} \cdot (\sqrt{\frac{r-m}{m}} + \sqrt{\frac{m}{r-m}} - \frac{\lambda \cdot r}{\sqrt{(r-m)m}}) - \epsilon$$

$$= \frac{(1-\lambda)r}{2\sqrt{(r-m)m} + r} - \epsilon$$

Let $\alpha = m/r \in (0,1)$ denote for the proportion of chosen classes with respect to all classes, we find that $C = (1 + 2\sqrt{(1-\alpha)\alpha})^{-1}(1-\lambda)$. Thus, the extremum is taken at $\alpha = 1/2$, and we have $C \geq \frac{1}{2} \cdot (1-\lambda)$, which is our theorem. □

Theorem 2 shows that when combining random greedy algorithm and double continuous greedy algorithm, and for $0 < \lambda < 1/(e-1)$, we could reach a 0.356-approximation. It could be easily shown by comparing the two bounds that when $m < 0.082r$ or $m > 0.918r$ we choose random greedy algorithm and when $0.082r \leq m \leq 0.918r$ we choose double continuous greedy algorithm.

## 2. Details of Continuous Double Greedy Algorithm

### 2.1. Reduction

To simplify our discussion in the original paper, we assume the following reduction of the original problem [2] is applied:

**Reduction 1.** *For the problem of $\max \{h(U) : |U| = m, U \subset B_u\}$, we may assume $2m < |B_u|$.*

*Proof.* If this is not the case, let $\bar{m} = |B_u| - m$ and $\bar{h}(U) = h(B_u \backslash U)$, it could be easily checked that $2\bar{m} < |B_u|$ and the problem $\max \{\bar{h}(U) : |U| = \bar{m}, U \subset B_u\}$ is equivalent to the original problem. □

The details of Algorithm 4 in the original paper are based on the assumption $2m \leq |B_u|$.

### 2.2. Initial State of Dynamic Programming

The explanation for $P(s_{n,[j]} \geq q_{n,[i]})$ is the probability of the $j$th-largest similarity between base classes in the random set $S \in B_s$ and the novel class $n$ larger than $q_{n,[i]}$, i.e. the $i$th-largest similarity between base classes in $B_u$ and the novel class $n$. From this definition, the initial state of the dynamic programming process is:

$$P(s_{n,[1]} \geq q_{n,[1]}) = x_{[1]}$$

$$P(s_{n,[j]} \geq q_{n,[1]}) = 0 \quad for \quad j = 2, 3, \cdots K$$

$$P(s_{n,[1]} \geq q_{n,[i]}) = (1 - P(s_{n,[1]} \geq q_{n,[i-1]})) \cdot x_{[i]} \quad for \quad i = 2, 3, \cdots |B|$$

## 2.3. Pruning of Dynamic Programming

According to Equation 6 and 7 in original paper, we need to calculate $P_u(s_{n,[j]} \geq q_{n,[i]})$ for $j = 1 \cdots K$ and $i = 1 \cdots |B|$, for each $u \in B_u$ and novel class $n$. Noticing that here we use $P_u$ instead of $P$ because for each $u \in B_u$, we must set $x_u = (x \wedge (B_u - u))_u = 0$ and run dynamic programming by Equation 7. Thus the result for $P(s_{n,[j]} \geq q_{n,[i]})$ is different considering selecting different $u$. Traditionally, we need to fix and loop $u \in B_u$, $n \in N$ to calculate $P_u(s_{n,[j]} \geq q_{n,[i]})$ in $O(K \cdot |B|^2 \cdot |N|)$. However, in this section, we introduce a pruning method, which could largely decrease the time complexity.

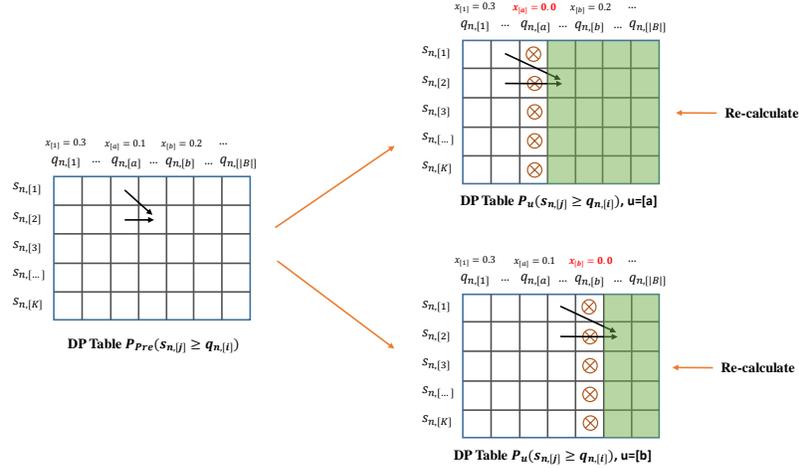

Figure 1. Example of Pruning Methods

The keypoint is that we could pre-compute $P_{pre}(s_{n,[j]} \geq q_{n,[i]})$, as Figure 1 shows. The dynamic programming (DP) table of $P_{pre}(s_{n,[j]} \geq q_{n,[i]})$ is constructed by setting the probability vector $x$ as its original value, without setting certain $x_u$ to be 0. In this way, when different $u \in B_u$ is selected, we could utilize this pre-calculation DP table. The unique difference for calculating $P_{pre}(s_{n,[j]} \geq q_{n,[i]})$ and $P_u(s_{n,[j]} \geq q_{n,[i]})$ is that we need to set corresponding $x_u = 0$, as the right part of Figure 1 shows. Let us suppose $u = [a]$ and we encourage two pruning methods in this paper: First, if $|P_{pre}(s_{n,[j]} \geq q_{n,[a]}) - P_{pre}(s_{n,[j]} \geq q_{n,[a-1]})| < \epsilon$ for all $j = 1, \cdots, K$, then there is no need to re-calculate $P_u(s_{n,[j]} \geq q_{n,[i]})$, and we could directly use $P_{pre}(s_{n,[j]} \geq q_{n,[i]})$ instead. Noticing that when $a$ is relatively large, there is a high probability satisfying the condition above, thus we could decrease the constant number of the time complexity of the algorithm substantially. Second, even though there is need to re-calculate DP table for $P_u(s_{n,[j]} \geq q_{n,[i]})$, we find that the left part of the DP table of column $a$ does not need to re-calculate as well, as Figure 1 shows. In this way, we could only re-calculate the right part (the green area). By using these two pruning methods simultaneously, the general complexity of the algorithm is relatively low compared with the worst case $O(T \cdot K \cdot |B|^2 \cdot |N|)$. The algorithm could further be easily extended to parallel computing version for a greater acceleration.

## 2.4. Pipage Rounding

The original paper mentions that, to transform the fractional solution obtained by Algorithm 4 to an integral solution, we may use some rounding techniques. One of the classical trick is Pipage Rounding.

We need three things to make Pipage Rounding work:

1. For any $x \in \mathcal{P}$, we need a vector $v$ and $\alpha, \beta > 0$ such that $x + \alpha v \in \mathcal{P}$ or $x - \beta v \in \mathcal{P}$ have strictly more integral coordinates.

2. For all $x$, the function $g_x(t) := F(x + tv)$ needs to be convex.

3. Finally, we need a starting fractional $x$ with a guarantee that $F(x) \geq \rho \cdot OPT$.

where $\mathcal{P} = \{x \in [0,1]^{|B_u|} : \sum_{j=1}^{|B_u|} x_j = m\}$ is a polytope constraint and $F(\cdot)$ is the multi-linear extension of the original optimization function $h(\cdot)$.

Noticing that the assumption 2 and 3 are satisfied in Non-monotone Submodular Optimization, where assumption 2 is proved by [1], and assumption 3 is consistent with Theorem 4 in original paper. Next we focus on assumption 1.

Suppose $x$ is a non-integral vector in $\mathcal{P}$ and there are at least two fractional coordinates. Let it be $x_p$ and $x_q$. Define $v = e_p - e_q$, where $e_p$ is the vector with 1 in the $p$th coordinate and 0 elsewhere. Let $\alpha = min(1 - x_p, x_q)$ and $\beta = min(1 - x_q, x_p)$. After constructing $v, \alpha, \beta$, easy to show that $x + \alpha v$ and $x - \beta v$ are both in $\mathcal{P}$ and both of them have strictly more integral coordinates.

We show that all three assumptions are satisfied, for running Pipage Rounding, we select two coordinates of $x$ at each time, selecting $v, \alpha, \beta$ as above, compare $F(x + \alpha v)$ and $F(x - \beta v)$ and pick the probability vector making the value larger (*i.e.* $x + \alpha v$ or $x - \beta v$) as the new probability vector $x$. When calculating the value of the function, we still just need to calculate $F(x + \alpha v) - F(x)$ instead of directly calculating $F(x + \alpha v)$ by the equation:

$$F(\cdots, 1, \cdots, x'_q, \cdots) - F(\cdots, x_p, \cdots, x_q, \cdots) =$$
$$(F(\cdots, 1, \cdots, x'_q, \cdots) - F(\cdots, x_p, \cdots, x'_q, \cdots)) +$$
$$(F(\cdots, x_p, \cdots, x'_q, \cdots) - F(\cdots, x_p, \cdots, x_q, \cdots)),$$

and convert the problem of change in two coordinates to change in only one coordinate, which could be solved using dynamic programming the same as Equation 6 and 7 with a slight difference, as is the case of $F(x - \beta v)$. Repeat this process until the component of $x$ is all integral (*i.e.* 1 or 0). From assumption 1 we know that the algorithm will definitely converged to an integral solution. [1] also shows that the integral solution $x^*$ after running Pipage Rounding also satisfies $F(x^*) \geq F(x)$, which does not change the lower bound of Continuous Double Greedy Algorithm.

### 2.5. Extensions

We also note that Algorithm 4 (along with Algorithm 3) in original paper has more applicability in real cases, especially when there are some modular constraints. For example, we could add a constraint to the original problem that the difficulty of obtaining a sufficient image set for each base class could be quantified as a real number, and we should balance the accuracy of classification on novel classes and the total difficulty of obtaining base dataset when selecting base classes. The setting is equivalent to substact a hyper-parameter $\mu$ multiplying the total difficulty from the original optimization function. Noticing the total difficulty is a modular term, thus the new optimization function is also submodular and we could still solve this new problem by non-monotone submodular optimization.

## 3. Complexity Analysis

For Algorithm 1, we use a balanced binary search tree to record the similarity of base classes with each novel class. Establishing and updating the search tree cost $O(|B| \cdot log|B| \cdot |N|)$ totally.

For Algorithm 2 and 3, we use a minimum heap to record current top-$K$ similar base classes for each novel class. For each turn, the calculation of all $h(u_i|U_{i-1})$ costs $O(|B|)$, for Algorithm 2 finding the top-1 of $h(u_i|U_{i-1})$ costs $O(|B|)$ and for Algorithm 3 finding top-m elements costs $O(|B| \cdot logm)$. Finally the update of the minimum heap costs $O(|N| \cdot logK)$. Thus totally the complexity is $O(m \cdot (|B| + |N| \cdot logK))$ for Algorithm 2 and $O(m \cdot (|B| \cdot logm + |N| \cdot logK))$ for Algorithm 3.

For Algorithm 4, for each turn $t$ and for each $u \in B_u$, the dynamic programming process costs $O(K \cdot |B| \cdot |N|)$. Thus, the worst-case complexity of the Double Continuous Greedy Algorithm is $O(T \cdot K \cdot |B|^2 \cdot |N|)$. However, with some pruning strategy (see Appendix), the constant number of the complexity is relatively low (much smaller than 1).

## 4. More Ablation Studies

### 4.1. Effects of Model Head

Table 1. ImageNet: Pre-trained Selection, 100-way novel classes

| Algorithm | Head | m=100, 5-shot | m=100, 20-shot | m=20, 5-shot | m=20, 20-shot |
|---|---|---|---|---|---|
| Random | 1-NN | 39.39% ± 0.82% | 49.47% ± 0.67% | 23.89% ± 0.56% | 33.06% ± 0.47% |
|  | SR | 38.74% ± 0.76% | 50.20% ± 0.40% | 24.29% ± 0.49% | 36.38% ± 0.37% |
| DomSim | 1-NN | 38.00% ± 0.36% | 48.80% ± 0.79% | 23.15% ± 0.43% | 31.81% ± 0.58% |
|  | SR | 38.84% ± 0.74% | 52.81% ± 0.20% | 23.62% ± 0.29% | 36.31% ± 0.45% |
| Alg. 1, $K=1, \lambda=0$ | 1-NN | 43.42% ± 0.78% | 53.79% ± 0.37% | 25.71% ± 0.43% | 34.67% ± 0.36% |
|  | SR | **43.72%** ± **0.47%** | 55.84% ± 0.38% | 26.08% ± 0.45% | 37.75% ± 0.22% |
| Alg. 2, $K=1, \lambda=0$ | 1-NN | 43.20% ± 0.76% | 53.61% ± 0.27% | 26.13% ± 0.44% | 34.97% ± 0.45% |
|  | SR | 43.70% ± 0.56% | **55.87%** ± **0.46%** | **26.60%** ± **0.55%** | **38.28%** ± **0.28%** |
| Alg. 2, $K=3, \lambda=0$ | 1-NN | 42.89% ± 0.43% | 53.13% ± 0.27% | 25.10% ± 0.48% | 34.52% ± 0.51% |
|  | SR | 43.02% ± 0.11% | 55.74% ± 0.21% | 25.71% ± 0.41% | 37.68% ± 0.25% |

The goal of this section is to prove that our proposed algorithm is not influenced by the choice of few-shot learning algorithm. We try different model heads after extracting the high-level features of the backbone. We select the Pre-trained Selection setting on ImageNet to demonstrate the viewpoint. The result is shown in Table 1. 1-NN means that we use a 1-NN algorithm based on cosine similarity to give the label of a test sample as the one with the nearest class centroid. SR means Softmax Regression on the high-level representation space. (*i.e.* Fine-tuning the classification layer in original backbone). The two methods represent for an easy realization of metric-based method and learning-based method. From Table 1 we show that the promotion of SR compared with 1-NN for all selection algorithms is rather stable in the same experiment setting and our algorithm is model-agnostic. Moreover, comparing 5-shot with 20-shot, we find that when the shot number is increasing, the margin of our algorithm and the baselines is shrinking when using SR as model head, which shows that the effect of fine-tuning gradually surpasses the effect of class selection with the increase of the shot number, demonstrating that our algorithm performs much better on few-shot setting.

### 4.2. Effects of the Number of Novel Classes

Table 2. ImageNet: Pre-trained Selection, 10-way

| Algorithm | m=100, 20-shot |
| --- | --- |
| Random | $84.33\% \pm 1.71\%$ |
| DomSim | $85.78\% \pm 2.06\%$ |
| Alg. 1, $K = 10, \lambda = 0$ | $88.36\% \pm 1.15\%$ |
| Alg. 2, $K = 3, \lambda = 0$ | $87.62\% \pm 1.38\%$ |
| Alg. 2, $K = 10, \lambda = 0$ | $88.02\% \pm 1.58\%$ |
| Alg. 4, $K = 10, \lambda = 0.2$ | $\mathbf{88.52\% \pm 1.88\%}$ |

In this section, we show the experiment result for 10-way 20-shot setting with $m = 100$ with 1-NN head in Table 2. We could draw three conclusions: First, in 10-way setting, our algorithm promotes about $4.19\%$ compared with Random Selection, which is at the same level with 100-way 20-shot setting shown in Table 1, demonstrating the effectiveness of our proposed algorithm in different number of novel classes. Second, we find that we need to increase $K$ compared with 100-way 20-shot setting as the number of base classes far exceeds the number of novel classes, thus we could provide more similar base classes for each novel class to improve the performance. Third, compared with $\lambda > 0$ and $\lambda = 0$ case, we show that diversity may be helpful when the number of base classes is much larger than the number of novel classes. In this setting diversity brings about a promotion of $0.5\%$.

### 4.3. Cold Start Selection on Caltech and CUB dataset

Table 3. Cold Start Selection, 100-way

| Algorithm | m=100, 5-shot, Caltech | m=100, 5-shot, CUB |
| --- | --- | --- |
| Random | $18.46\% \pm 1.19\%$ | $45.31\% \pm 1.32\%$ |
| DomSim | $27.32\% \pm 0.82\%$ | $51.72\% \pm 1.24\%$ |
| Alg. 2, $K = 1, \lambda = 0$ | $27.59\% \pm 0.76\%$ | $53.48\% \pm 1.19\%$ |
| Alg. 2, $K = 3, \lambda = 0$ | $\mathbf{29.33\% \pm 0.69\%}$ | $\mathbf{53.56\% \pm 1.34\%}$ |
| Alg. 2, $K = 5, \lambda = 0$ | $28.83\% \pm 0.60\%$ | $53.33\% \pm 1.18\%$ |
| Pre-trained (Upperbound) | $29.65\% \pm 0.82\%$ | $55.41\% \pm 1.25\%$ |

In this section, we also test cold start selection on Caltech256 and CUB-200-2011 as Table 3. All our algorithms use a mechanism of 6-12-25-50-100. The conclusion is the same as the original paper and there is nothing to discuss more about the results.

## 5. Detailed Experiment Settings in Training Phase

For all experiments, when training the base model, we use a standard ResNet-18 structure. The output dimension of the high-level feature is 512. The preprocessing step of the images is the same as original ResNet-18 paper. The base model is trained for 120 epoches, the learning rate is set to 0.1 for Epoch 1 to Epoch 25, 0.01 for Epoch 25 to Epoch 50, 0.001 for Epoch 50 to Epoch 80, 0.0001 for Epoch 80 to Epoch 105 and 0.00001 for Epoch 105 to Epoch 120. A weight decay with hyperparameter 0.0005 is used. We use a momentum SGD and the momentum coefficient is set to 0.9. The batch size is set to 64. We train the whole base model on 8*Nvidia Tesla V100. For each base model, the training time is about 4 hours and for

each experiment setting this training process is repeated for 10 times, the total training hours for each experiment setting (*i.e.* each result number in the result tables) is about 40 hours. (The cold start problem may spend much longer time, about 75 hours per experiment setting). The main framework of the training process is based on Tensorflow, and the selection algorithm is based on C++11 for speed-up.



\end{document}

%% file: article.tex
\vspace{-1.0em}
\section{Introduction}

Few-shot Learning \cite{miller2000learning, fei2006one} is a branch of Transfer Learning, its basic setting is to train a base model on the base dataset consisting of base classes with ample labeled samples, then adapt the model to a novel support set consisting of novel classes with few samples, and finally evaluate the model on the novel testing set consisting of the same novel classes as the novel support set.


Traditionally, many works focus on how to learn meta-knowledge from a fixed base dataset. The generation process of the base datasets generally depends on random selection or human experience, which is not necessarily perfect for few-shot learning. Due to the fact that the fine-tuning mechanism on the novel support set is not as effective as learning with large-scaled training samples on novel classes \cite{vinyals2016matching}, the base dataset plays a critical role for the performance of few shot learning. Till now, however, we have little knowledge on how to measure the quality of a base dataset, and not to mention how to optimize the its selection process.

The targeting problem described above is somewhat related to Curriculum Learning \cite{Bengio2009Curriculum, tsvetkov2016learning} and data selection in transfer learning \cite{remus2012domain, ruder2017learning, qu2019learning}. Different from Curriculum Learning aiming to speed up learning of provided classes, we focus on learning to select base classes in a transfer learning manner, where the selected base classes are used for classification on novel classes. With respect to the data selection methods in transfer learning, first, our problem is a class-based selection instead of sample-based selection problem, which significantly decreases the search space for selection. Second, we consider the problem in a few-shot learning scenario, where there is no validation dataset on novel classes, and modern methods with feedback mechanism on validation performance (\textit{e.g.} Bayesian Optimization in \cite{ruder2017learning}, Reinforcement Learning in \cite{qu2019learning}) are not applicable.

Here we consider a realistic and practical setting that $M$ base classes are to be selected from $N$ candidate classes, and each candidate class contains only a small number of labeled samples before selection. Once the $M$ classes are selected, one could expand the samples of these selected classes to a sufficient size by manually labeling, which are further used to construct the base dataset and train the base model. The selection process could be conducted either in an one-time or incremental manner.

To solve the problem, we confront two challenges. First, the problem is a discrete optimization problem. The complexity of naive enumeration method is $O(N^M)$, which is intractable in real cases. Second, there is no touchable way to optimize the classification performance of novel classes directly, hence we need to find a proxy indicator that is both easy to optimize and highly correlated with the classification performance on novel classes.

In this paper, we find a simple yet effective indicator Similarity Ratio, first proposed by our previous work \cite{zhou2019learning}. For a candidate class, the Similarity Ratio considers both its similarities with novel classes and diversity in base classes. We demonstrate that this indicator is highly and positively correlated with the performance of few-shot learning on the novel testing set. We theoretically prove that this indicator satisfies submodular property, which pledges us to obtain a sub-optimal solution in polynomial time complexity. Thus, the base class selection problem could be surrogated by optimizing a variant of Similarity Ratio. We carry out extensive experiments on three different cases: the Pre-trained Selection, the Cold Start Selection, and the General Selection on ImageNet, Caltech256, and CUB-200-2011 datasets. Results show that our method could significantly improve the performance of few-shot learning in both general image classification and fine-grained image classification. The performance improvement margin is rather stable regardless of the distribution transfer from the support set to the query set, change of few-shot model, or change of few-shot experimental settings.

\vspace{-0.3em}
\section{Related Work}
\noindent \textbf{Few-shot Learning} The concept of One-shot Learning is proposed by \cite{fei2006one}, and a more general concept is Few-shot Learning.  Three mainstreams of approaches are identified in the literature. The first group is based on a meta-learning manner, including Matching Network \cite{vinyals2016matching}, MAML\cite{finn2017model}, Prototypical Network \cite{snell2017prototypical}, Relation Network \cite{sung2018learning}, SNAIL \cite{mishra2017simple} \textit{etc}, which learn an end-to-end task-related model on the base dataset that could generalize across all tasks. The second group of methods is learning to learn image classifiers for unseen categories via some transfer mechanism while keeping the representation space unchanged. The advantage of these methods is to avoid drastically re-training the model and more friendly to extremely large base datasets and model, \textit{e.g.} classification on ImageNet. Common methods are MRN \cite{wang2016learning}, CLEAR \cite{kozerawski2018clear}, Weight Imprinting \cite{qi2018low}, VAGER \cite{zhou2019learning} \textit{etc.} The third group of methods is to apply data generation. The core idea is to use a pre-defined form of generation function to expand the training data of unseen categories. Typical work includes \cite{hariharan2017low} and \cite{wang2018low}.

\noindent \textbf{Data Selection} The underlying assumption of data selection is that not all training data is helpful to the learning process; some training data may even perform negative effects. Thus, it's important to distinguish \textit{good} data points from \textit{bad} data points to improve both the convergence speed and the performance of the model. Roughly there are two branches of work: one is to assume training data and testing data are sampled from the same distribution, a common way to deal with this problem is to reweight the training samples \cite{kumar2010self, tsvetkov2016learning, fan2017learning}, which is out of the scope and will not be covered in this paper. The other branch is data selection in a transfer learning manner. Mainstream approaches include that \cite{remus2012domain} proposes a method based on heuristically defined distance metric to find most related data points in the source domain to the target domain; \cite{ruder2017learning} views the effect of data selection process to final performance of the classification on target domain as a black box model and uses Bayesian Optimization to iteratively adjust the selection through performance on validation dataset and further \cite{qu2019learning} substitutes Bayesian Optimization to Reinforcement Learning, which is more suitable to introduce deep model to encourage more flexibility in designing selection algorithms.

\vspace{-0.3em}
\section{Preliminary Study}
\vspace{-0.3em}
\subsection{Similarity Ratio}
\cite{zhou2019learning} first proposes a concept called Similarity Ratio (SR) defined for each novel class as:
\begin{equation}
    \mbox{SR} = \frac{\mbox{Average Top-K Similarity with Base Classes}}{\mbox{Average Similarity with Base Classes}}.
\end{equation}
Here the similarity of two classes is determined by a specific metric on the representation space, \textit{e.g.} the cosine distance of two class centroids. Among all base classes, we sort the similarity of each base class with the corresponding novel class in a descent order. The numerator is calculated by averaging the similarity of the top-K similar base classes and the denominator is calculated by averaging the similarity of all base classes. To improve SR, the numerator indicates there should be some similar base classes with the corresponding novel class and the denominator indicates the base classes should be diversified conditioned on each novel class. \cite{zhou2019learning} further points out that the few-shot performance is positively correlated with this indicator.

\vspace{-0.3em}
\subsection{The Relationship Between SR and Few-shot Learning Performance}
In this part, we will show more evidence from a statistical perspective of the relationship between SR and few-shot learning performance.

Specifically, a preliminary experiment is conducted as follows: we randomly choose 500 classes from ImageNet dataset, and further split them into 400 base classes and 100 novel classes. For each few-shot classification setting, we randomly select 100 base classes over 400 as the base dataset, and using all 100 novel classes to perform a 100-way 5-shot classification. A ResNet-18 \cite{he2016deep} is trained on the base dataset, and we extract the high-level image features (512-dimensional features after conv5\_x layer) for novel support set and novel testing set. We calculate the average feature for each novel class in the novel support set as the class centroid and directly use 1-nearest neighbor based on the cosine distance metric defined on the representation space to obtain the Top-1 accuracy for each novel class of the testing set. The base dataset selection, training and evaluating process is repeated for 100 times and for each novel class, we run the regression model:
\begin{equation}
\label{eqn:reg}
    Acc = \beta_1 \cdot x_1 + \beta_2 \cdot x_2 + \alpha + \epsilon
\end{equation}
\[
    \left\{\begin{aligned}
        x_1 &= \mbox{Average Top-K Similarity with Base Classes}\\
        x_2 &= \mbox{Average Similarity with Base Classes}
    \end{aligned}\right.
\]
where $Acc$ represents for the Top-1 accuracy for the corresponding novel class, $\alpha$ represents for the residual term and $\epsilon$ represents for noise. The similarity of two classes in this regression model is calculated by the cosine distance of two centroids defined on the representation space of ResNet-18 trained by all 400 candidate base classes. Hence, totally we could obtain 100 regression models, each for a novel class, and each model is learned under 100 data points related to 100 different choices of base dataset.

\begin{figure}[ht]
    \centering
    \includegraphics[width=2.2in]{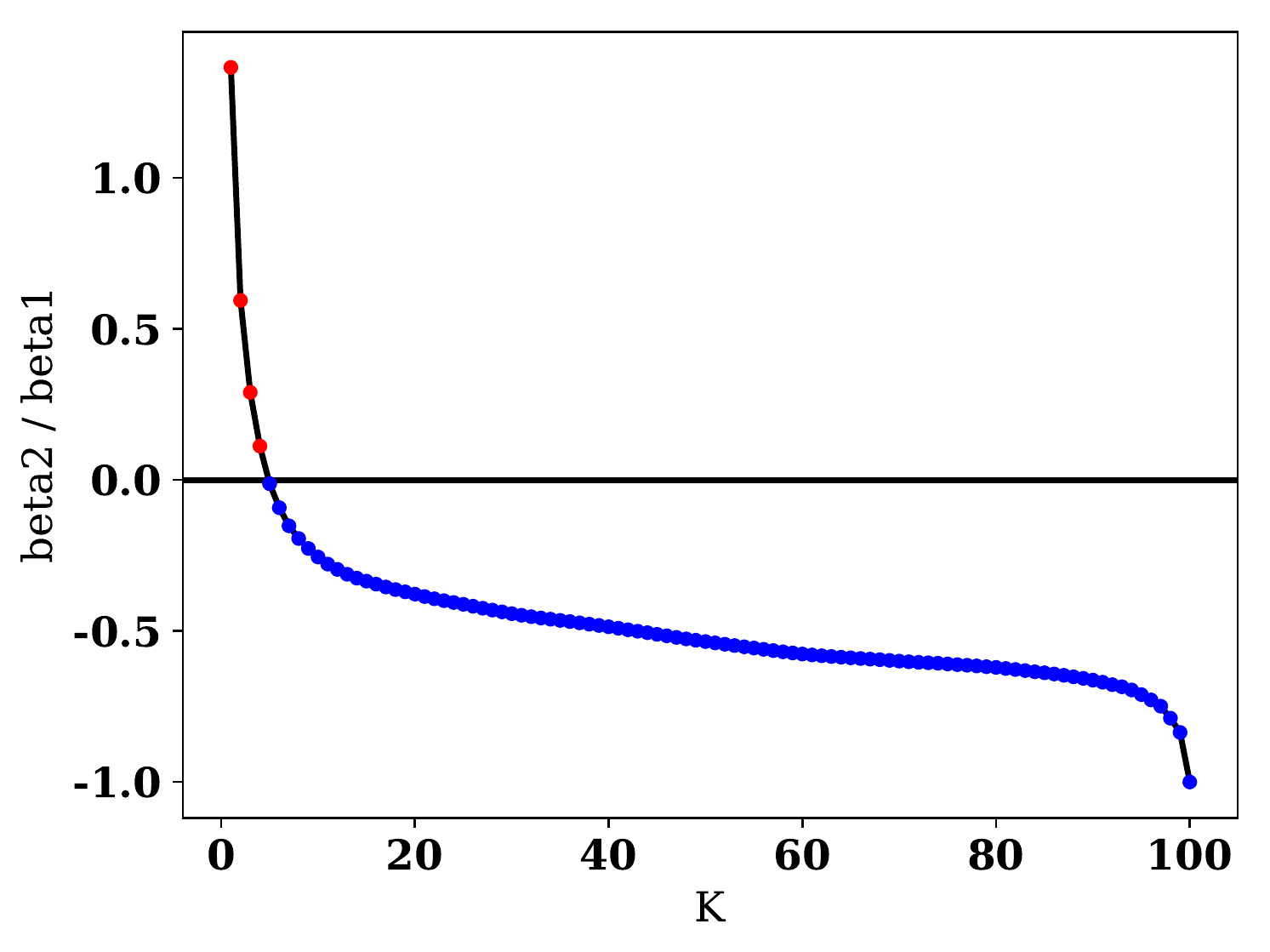}
    \caption{The coefficient $\bar{\beta_2} / \bar{\beta_1}$ changed with K.}
    \label{fig:coefficient}
\end{figure}

\begin{figure}[htbp]
    \centering
    \subfigure{
        \begin{minipage}[t]{0.45\linewidth}
        \centering
        \includegraphics[width=1.4in]{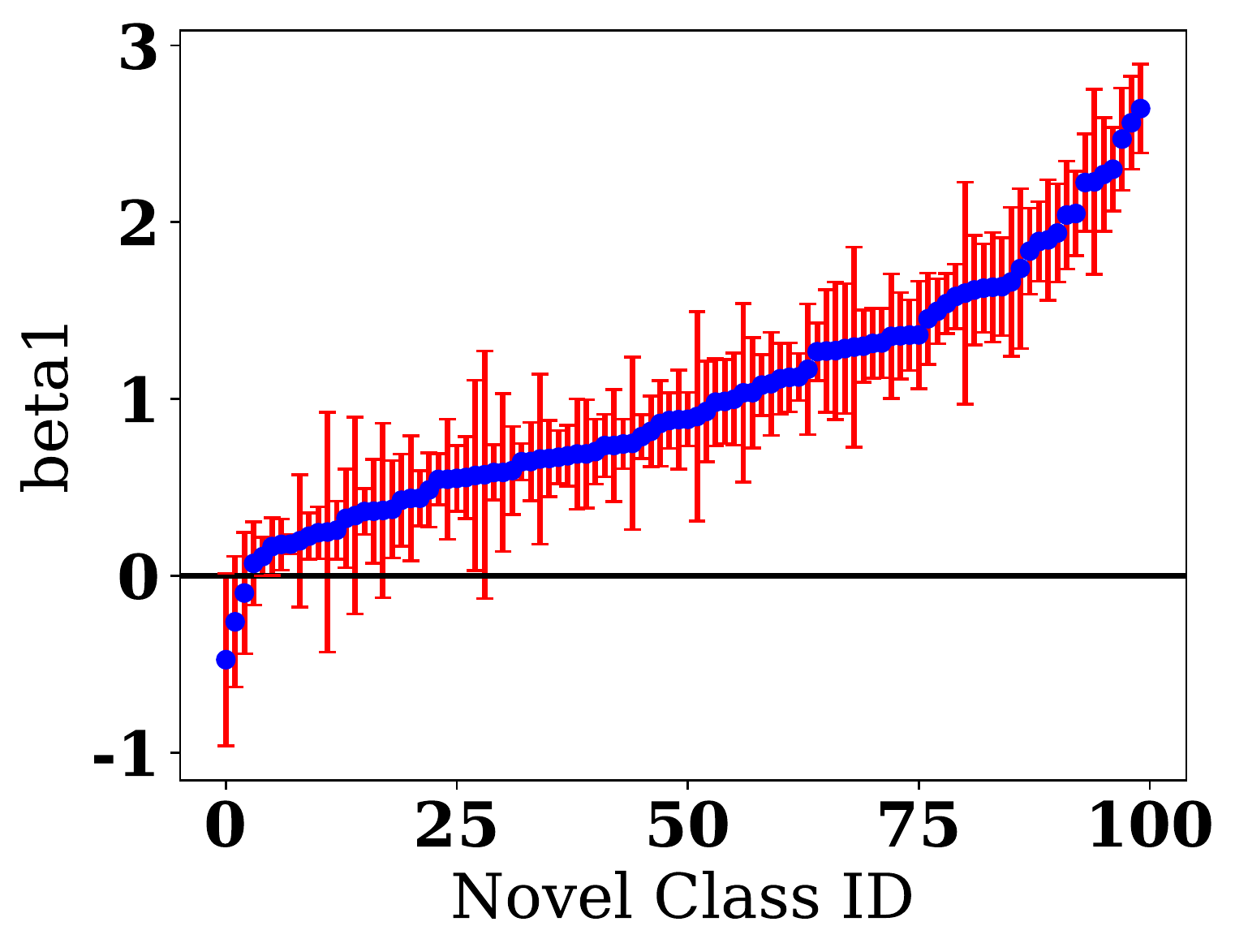}
        \end{minipage}
    }
    \subfigure{
        \begin{minipage}[t]{0.45\linewidth}
        \centering
        \includegraphics[width=1.4in]{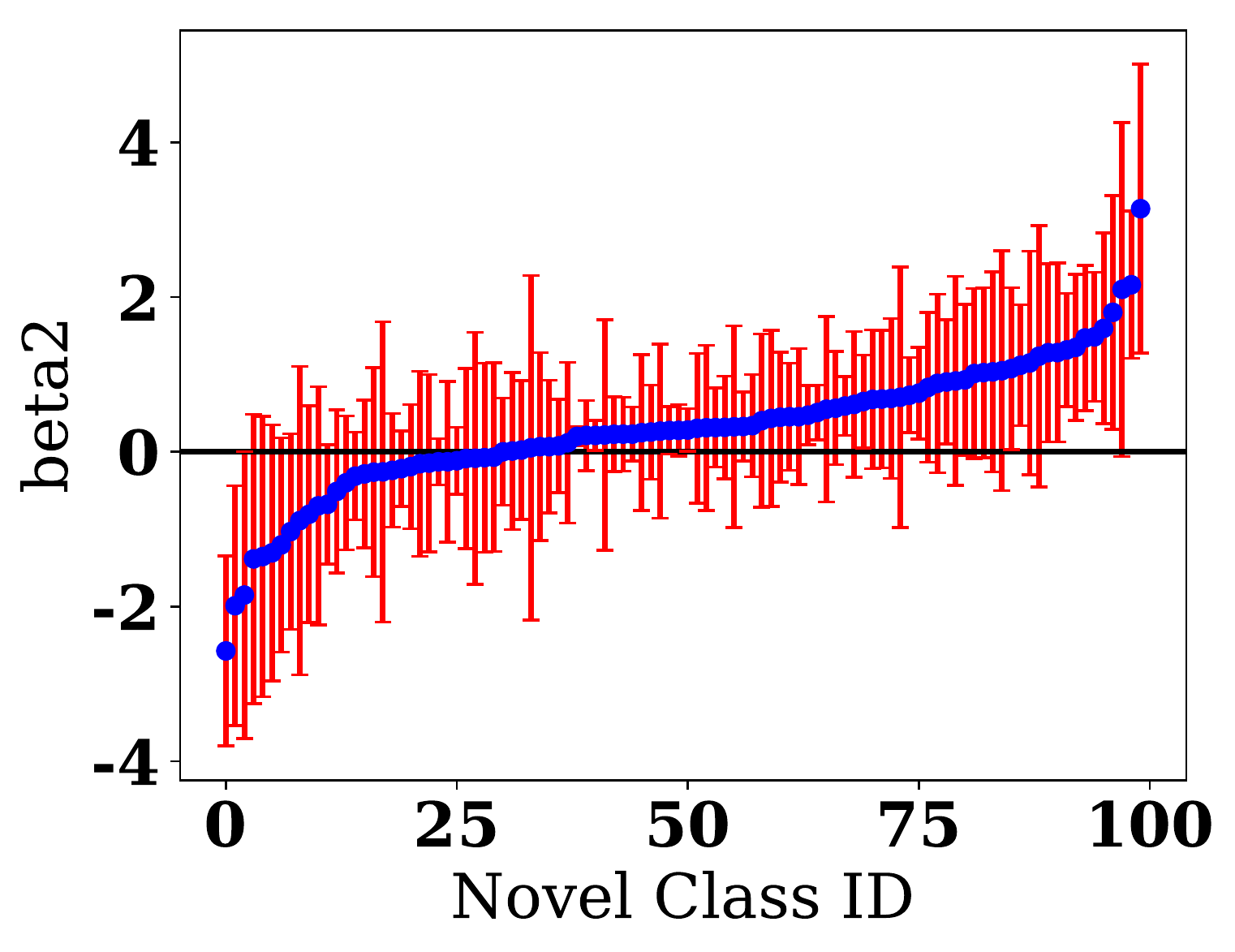}
        \end{minipage}
    }
    \subfigure{
        \begin{minipage}[t]{0.45\linewidth}
        \centering
        \includegraphics[width=1.4in]{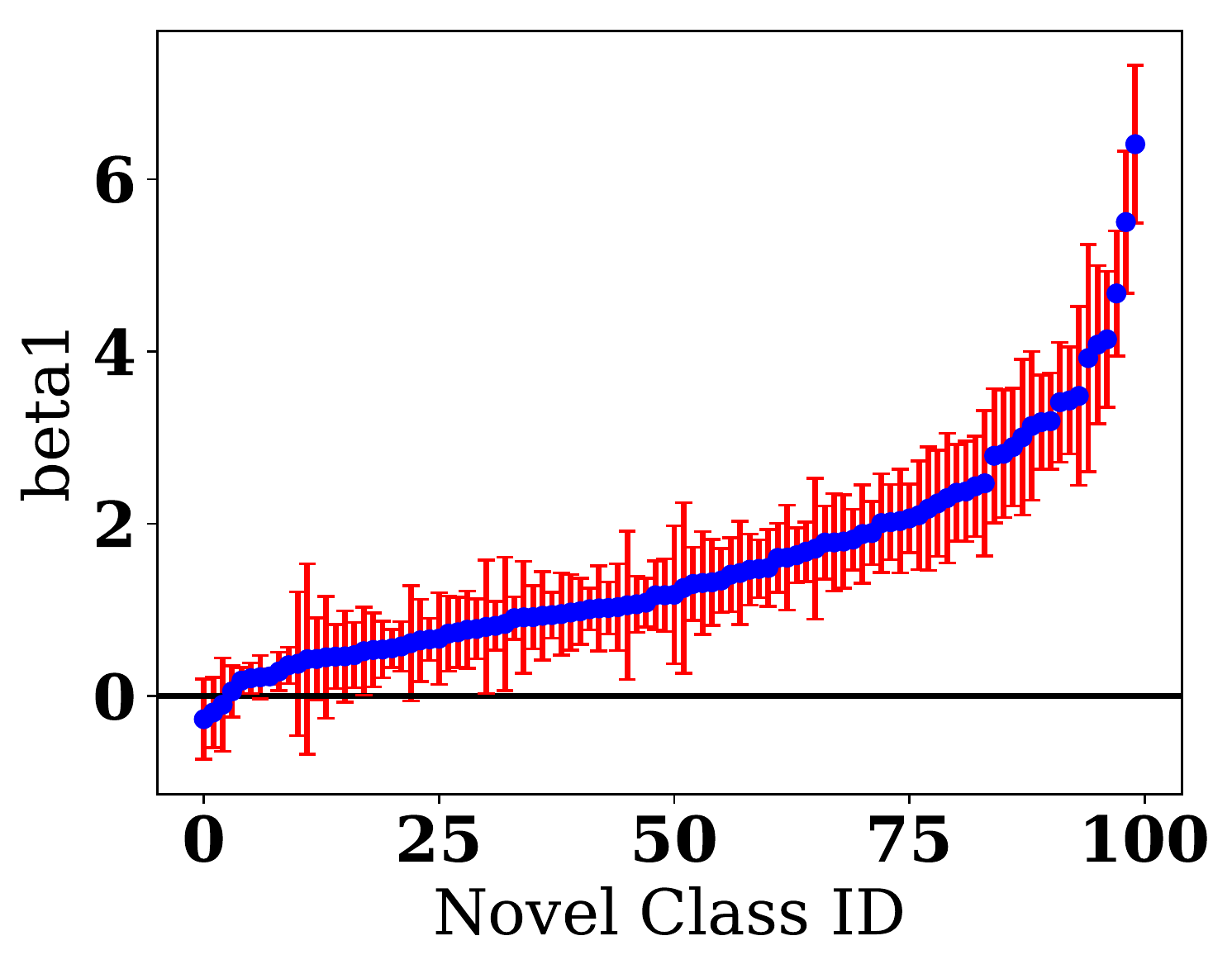}
        \end{minipage}
    }
    \subfigure{
        \begin{minipage}[t]{0.45\linewidth}
        \centering
        \includegraphics[width=1.4in]{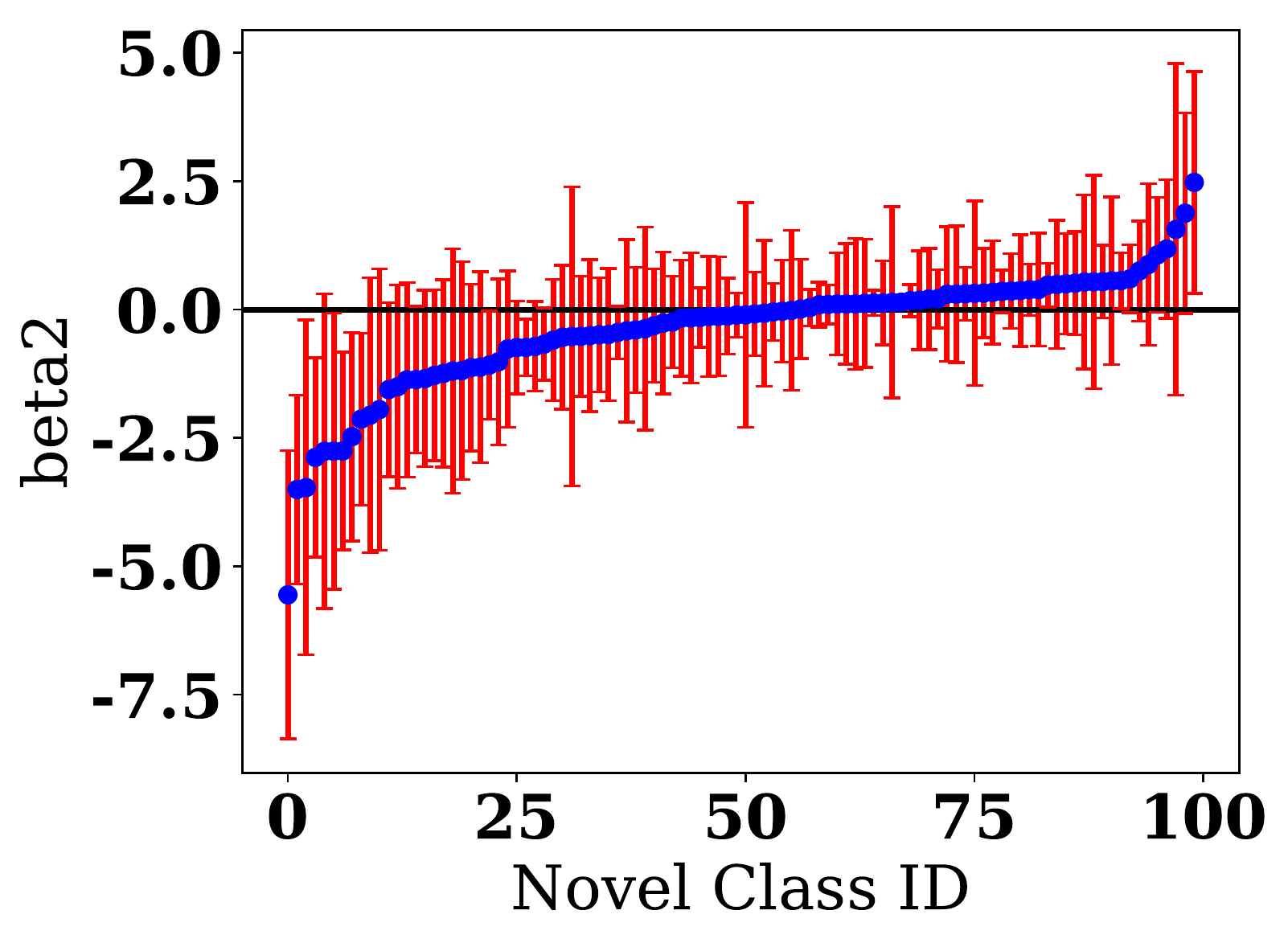}
        \end{minipage}
    }
    \centering
    \caption{We plot the coefficients $\beta_1$, $\beta_2$ of each novel class after sorting increasingly. The red bar represents for the $95\%$ confidence interval and the blue dot shows the exact coefficients. Top: result for Regression with $K=3$, $\bar{\beta_1}=0.99$, $\bar{\beta_2}=0.29$; Bottom: result for Regression with $K=10$, $\bar{\beta_1}=1.52$, $\bar{\beta_2}=-0.39$.}
    \label{fig:beta}
\end{figure}

\begin{figure}[ht]
    \centering
    \includegraphics[width=2.2in]{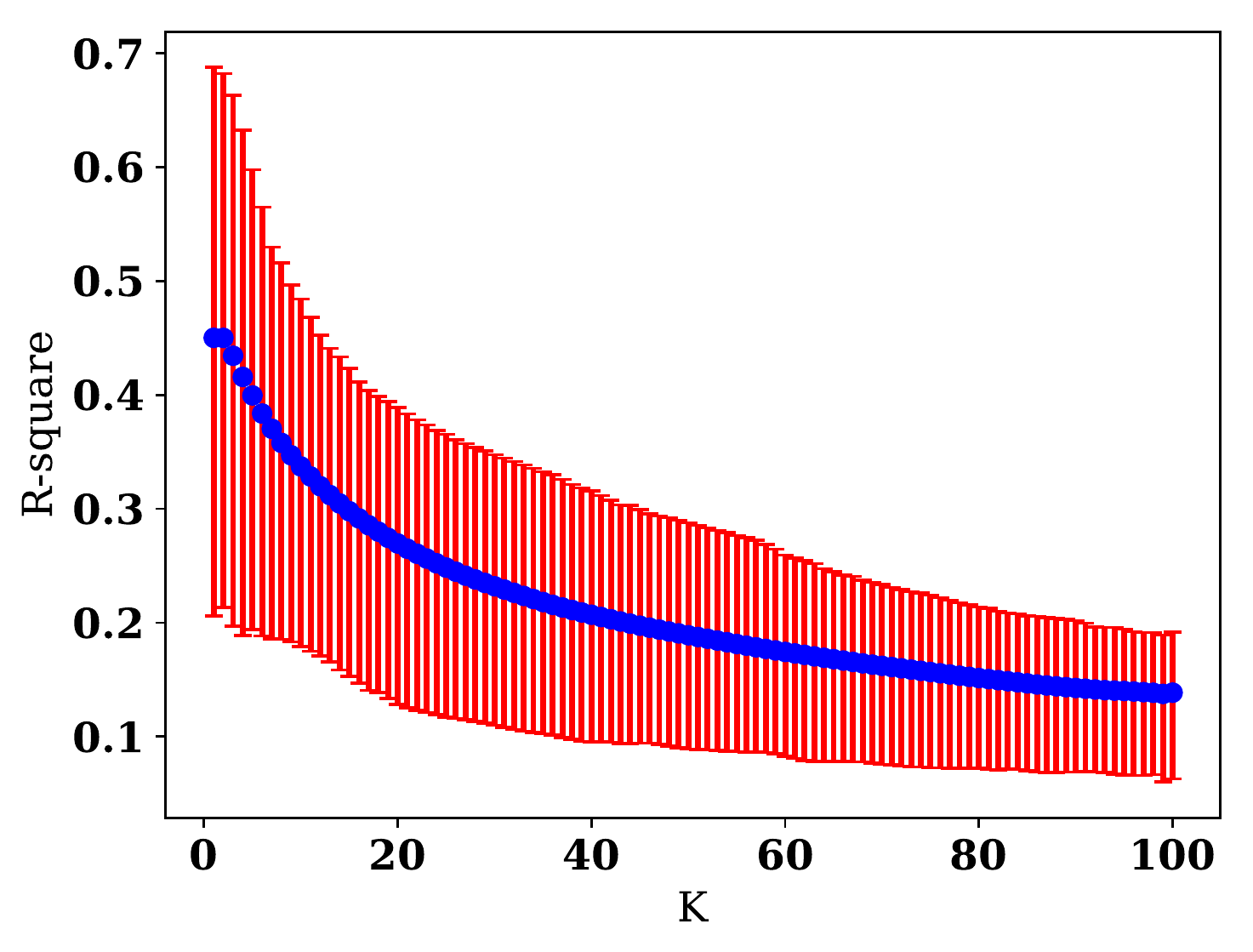}
    \caption{$R^2$ with the change of K for 100 regression models, the red bar represents for the interval from 25-quantile to 75-quantile, and the blue dot represents for the average $R^2$.}
    \label{fig:rsquare}
\end{figure}

With a different choice of $K$, the regression model may show different properties. We conclude our findings from Figure \ref{fig:coefficient}, \ref{fig:beta}, \ref{fig:rsquare}. 

We calculate the average of $\beta_1$ and $\beta_2$ for all novel classes, denoted as $\bar{\beta_1}$ and  $\bar{\beta_2}$. $\bar{\beta_1}$ is constantly positive in all choices of $K$, demonstrating the positive effect of Average Top-K Similarity to accuracy. Figure \ref{fig:coefficient} shows the change of coefficient $\bar{\beta_2} / \bar{\beta_1}$ with $K$. The result shows that $K=5$ is a demarcation point in this specific setting. The positive effect of Average Similarity (\textit{i.e.} $x_2$) will become negative after $K=5$. The reason is that when $K$ is small, the positive classes are insufficient, there is need to add more positive classes to improve the performance, and with the increase of $K$, the positive classes tend to saturate and there is an increasing need of negative classes to enhance diversity. In later main experiments, we set K to be a hyper-parameter.
    
Figure \ref{fig:beta} is a snapshot for the two settings with $K=3$ and $K=10$, which further proves the viewpoint above. Moreover, Figure \ref{fig:beta} gives more information about the distribution of $\beta_1$ and $\beta_2$.
    
Figure \ref{fig:rsquare} shows that the two components of the SR are relatively good proxy of the performance for few-shot learning when K is a small number (\textit{i.e.} The average $R^2$ reaches above 0.3 when $K \leq 10$). When $K=1$ the two components of SR explain about $45\%$ of the dependent variable.

Based on our findings, an optimization process could be designed to select core base classes.

\vspace{-0.3em}
\section{Algorithm}
\vspace{-0.3em}
\subsection{A Brief Introduction to Submodularity}

\begin{definition}
    Given a finite set $V = \{1, 2, \cdots, n\}$, a set function $f : 2^V \rightarrow \mathbb{R}$ is submodular if for every $A, B \in V$: $f(A \cap B) + f(A \cup B) \leq f(A) + f(B)$.
\end{definition}

A better way to understand submodularity property is that of diminishing returns: denote $f(u|A)$ as $f(A \cup {u}) - f(A)$, then we have $f(u|A) \geq f(u|B)$ for every $A \subseteq B \subseteq V$ and $u \notin B$. These two definitions are proved to be equivalent \cite{nemhauser1978analysis}. It has been proved that maximizing a submodular objective function $f(\cdot)$ is an NP-hard problem. However, with polynominal time complexity, several algorithms have been proposed to obtain a sub-optimal solution.

A function is monotone non-decreasing if $\forall A \subseteq B, f(A) \leq f(B)$. $f(\cdot)$ is called normalized if $f(\emptyset) = 0$. 

In this paper we mainly introduce a submodular optimization setting with cardinality constraint. The problem is formulated as: $max_{S \subseteq V, |S| = k} f(S)$, where $f(\cdot)$ is a submodular function. \cite{nemhauser1978analysis} shows that a simple greedy algorithm could be used to maximize a normalized monotone non-decreasing submodular fuction with cardinality constraints, with a worst-case approximation factor of $1 - 1/e \approx 0.632$. \cite{buchbinder2014submodular} shows that a normalized submodular function (may not be monotone non-decreasing) with an exact cardinality constraint $|S| = k$ could reach an approximation of $max \{\frac{1-k/en}{e}-\epsilon, (1 + \frac{n}{2\sqrt{(n-k)}k})^{-1} - o(1) \}$ with a combination of random greedy algorithm and continuous double greedy algorithm, where $k$ is the exact number of chosen elements and $n$ is the total number of elements. The proposed algorithm guarantees a 0.356-approximation, which is smaller than 0.632.

\vspace{-0.3em}
\subsection{Formulation}
Let $B_u$ represent for collection of unselected base classes, $B_s$ for selected base classes and $N$ for novel classes. The selection process is to select a subset $U$ with $m$ elements from $B_u$ and the base dataset is composed of $U$ and $B_s$. For each class $l$, we denote $c_l$ as certain class feature (\textit{e.g. its centroid of high-level feature}), and for each class set $A$, we denote $c_A = [c_{l_1}, c_{l_2}, \cdots c_{l_{|A|}}], l_1, l_2 \cdots l_{|A|} \in A$ as a collection of class features.

Next, we define an operator \textit{max-k-sum} as follows:
\[
    M^k(y) := \max \limits_{|K|=k}\sum_{i \in K}y_i = \sum^k_{j=1}y_{[j]},
\]
where $y$ is a numerical vector, $y_{[1]}, \cdots, y_{[n]}$ are the $y_i$'s listed in nonincreasing order. Based on our findings that SR is highly and positively correlated to the performance on novel classes in Section 3, the base class selection problem could be formulated as an optimization process on SR as a proxy. Concretely we have:
\begin{equation}
\label{eq:opt}
    \begin{split}
    \max \limits_{U \subset B_u \atop |U|=m} & \frac{1}{|N|} \sum_{n \in N} \frac{1}{K} \cdot M^K(f(c_n, \{c_{B_s}, c_U\})) \\
    & - \frac{\lambda}{|N|} \cdot \sum_{n \in N} \frac{1}{|B_s| + m} \sum_{u \in B_s \cup U} f(c_n, c_u),
    \end{split}
\end{equation}
where $f(c_a, \{c_{b1},\cdots, c_{bn}\}) = [f(c_a, c_{b1}),\cdots, f(c_a, c_{bn})]$ is a similarity function (\textit{e.g.} Cosine Distance). The optimization function is the same form of Equation \ref{eqn:reg}, where the first term is the numerator of SR and the second term is the denominator). $\lambda$ is seen as a hyper-parameter, whose meaning is equivalent to $-\bar{\beta_2} / \bar{\beta_1}$ in Section 3.2. $K$ is also a hyper-parameter. For simplicity we may assume $\lambda \geq 0$, as when $\lambda < 0$ the two terms of optimization function \ref{eq:opt} has a strong positive correlation, experiment results show there is not much improvement compared with directly setting $\lambda = 0$. $|U| = m$ is the cardinality constraint that exact $m$ base classes are needed to be selected.

The next corollary shows that Problem \ref{eq:opt} is equivalent to a submodular optimization.

\begin{corollary}
\label{cor:submodular}
    Considering optimization problem \ref{eq:opt}, when $\lambda = 0$, Problem \ref{eq:opt} is equivalent to a submodular monotone non-decreasing optimization with exact cardinality constraint and when $\lambda > 0$, Problem \ref{eq:opt} is equivalent to a submodular optimization with exact cardinality constraint.
\end{corollary}

\vspace{-0.3em}
\subsection{Optimization}

\vspace{-0.3em}
\subsubsection{Case 1: $\lambda = 0$}
The case $\lambda = 0$ could be seen as a standard submodular monotone non-decreasing optimization, hence we could directly use a greedy method on the value of target function, as Algorithm \ref{alg:B} shows. However, for this specific target function, a trivial setting with $m \geq K \cdot |N|$ needs further consideration. For this setting, a greedy algorithm on novel class (Algorithm \ref{alg:A}) could be proved to reach an optimal solution, while Algorithm \ref{alg:B} could just reach sub-optimal. Thus, the two different greedy algorithms are proposed to deal with the trivial and non-trivial case separately. For our description of the algorithms below, $f(\cdot, \cdot)$ denotes for the similarity function and $h(\cdot)$ denotes for the optimization function of Problem \ref{eq:opt} with $\lambda = 0$.

\begin{algorithm}
    \caption{Greedy Algorithm on Novel Class ($f, m$)}
    \begin{algorithmic}[1]
        \STATE {Let $U_0 \leftarrow \emptyset$, $S \leftarrow N$} 
        \FOR{$i=1$ to $m$}
            \STATE {Let $u \in B_u \backslash U_{i-1}$, $n \in S$ be the samples maximizing $f(c_u, c_n)$.}
            \STATE {Let $U_i \leftarrow U_{i-1} + u$, $S \leftarrow S - {n}$.}
            \IF {$S = \emptyset$}
                \STATE {$S \leftarrow N$.}
            \ENDIF
        \ENDFOR
        \RETURN {$U_m$}
    \end{algorithmic}
    \label{alg:A}
\end{algorithm}

\begin{algorithm}
    \caption{Greedy Algorithm on Target Function ($h$, $m$)}
    \begin{algorithmic}[1]
        \STATE {Let $U_0 \leftarrow \emptyset$} 
        \FOR{$i=1$ to $m$}
            \STATE {Let $u_i \in B_u \backslash U_{i-1}$ maximizing $h(u_i|U_{i-1})$.}
            \STATE {Let $U_i \leftarrow U_{i-1} + u_i$.}
        \ENDFOR
        \RETURN {$U_m$}
    \end{algorithmic}
    \label{alg:B}
\end{algorithm}

 We further give Thm. 1, 2 to show the optimization bound of the two algorithms. For this specific problem, the bounds are much tighter than the generic version in \cite{nemhauser1978analysis}.

\begin{theorem}
    For $B_s = \emptyset$ and $\lambda = 0$, when $m \geq K \cdot |N|$, using Algorithm \ref{alg:A} to solve for optimization problem \ref{eq:opt}, the solution will be optimal.
\end{theorem}

\begin{theorem}
    For $B_s = \emptyset$ and $\lambda = 0$, using Algorithm \ref{alg:B} to solve for optimization problem \ref{eq:opt}, let $h(\cdot)$ be the optimization function, and let Q be
    \[
        Q = \mathbb{E}_{u \sim Uniform(B), v \sim Uniform(N)}(f(c_u, c_v))
    \]
    representing for the average similarity between base classes and novel classes, we have $h(U) \geq (1- 1/e) \cdot h(OPT) + 1/e \cdot Q$, where $h(OPT)$ is the global optimal value of the optimization problem.
    \label{thm:2}
\end{theorem}

\vspace{-0.3em}
\subsubsection{Case 2: $\lambda > 0$}
The case $\lambda > 0$ could be seen as a non-monotone submodular optimization, with the technique in \cite{buchbinder2014submodular}, we combine both Random Greedy Algorithm (Algorithm \ref{alg:C}) and Continuous Double Greedy Algorithm (Algorithm \ref{alg:D}) for better optimization. The Random Greedy Algorithm is an extension of the standard Greedy Algorithm (Algorithm \ref{alg:B}), which is fit for settings with extremely low $m$. Details of the algorithm are given in Algorithm \ref{alg:C}.

\begin{algorithm}
    \caption{Random Greedy Algorithm ($h$, $m$)}
    \begin{algorithmic}[1]
        \STATE {Let $U_0 \leftarrow \emptyset$} 
        \FOR{$i=1$ to $m$}
            \STATE {Let $M_i \subset B_u \backslash U_{i-1}$ be a subset of size $m$ maximizing $\sum_{u \in M_i} h(u|U_{i-1})$.}
            \STATE {Let $u_i$ be a uniformly random sample from $M_i$.}
            \STATE {Let $U_i \leftarrow U_{i-1} + u_i$.}
        \ENDFOR
        \RETURN {$U_m$}
    \end{algorithmic}
    \label{alg:C}
\end{algorithm}

For much larger $m$, we will introduce the Continuous Double Greedy Algorithm. The core idea is to convert the discrete optimization of Problem \ref{eq:opt} to a continuous version.

Let $F(x)$ be the multilinear extension of the optimization function $h(\cdot)$ as:
\begin{equation}
    F(x) = \sum_{S \subseteq B_u}h(S) \prod_{u \in S}x_u \prod_{u \notin S}(1 - x_u)
\end{equation}
where $x \in [0,1]^{|B_u|}$. Given a vector $x$, $F(x)$ represents for the expectation of function $h$ given a random subset of $B_u$ with every element $u \in B_u$ \textit{i.i.d.} sampled with probability $x_u$ . For two vectors $x$ and $y$, define $x \vee y$ and $x \wedge y$ to be coordinate-wise maximum and minimum separately, \textit{i.e.} $(x \vee y)_u = \max(x_u, y_u)$ and $(x \wedge y)_u = \min(x_u, y_u)$. An important property for multilinear form function $F$ is:
\begin{equation}
    \frac{\partial F(x)}{\partial x_u} = F(x \vee u) - F(x \wedge (B_u - u))
\end{equation}
For simplicity, in this part, notation for a subset could also be represented as a 0-1 vector where the corresponding elements belonging to the subset are 1 and otherwise 0, consistent with \cite{buchbinder2014submodular}. In the double continuous greedy algorithm, we don't need to calculate the exact value for $F(x)$, the only difficulty is to calculate $F(x \vee u) - F(x \wedge (B_u - u))$. Theorem \ref{thm:3} gives a dynamic programming for fast calculation.

\begin{theorem}
    Let $S \subseteq B_u$ be a random set, with each element $v$ in $B_u$ \textit{i.i.d.} sampled with probability $(x \wedge (B_u - u))_v$. For each novel class $n \in N$, sort the similarity function $f(c_n, c_b)$ for each base class $b \in B = B_u \cup B_s$ in descent order, denoting as $q_{n, [1]}, q_{n, [2]}, \cdots q_{n, [|B|]}$, also, sort the similarity function for every base class in $S \cup B_s$ in descent order, denoting as $s_{n, [1]}, s_{n, [2]}, \cdots s_{n, [|S| + |B_s|]}$, then we have:
    \begin{footnotesize}
    \begin{align}
    \label{eqn:derivative}
    \begin{split}
        &F(x \vee u) - F(x \wedge (B_u - u)) \\
        & \!=\! \frac{1}{|N|\!\cdot\! K} \!\sum_{n \in N} \!\sum_{i=1}^{|B|} \!P(s_{n, [K]} \!=\! q_{n, [i]}) \!\max(f(c_n, c_u) \!-\! q_{n, [i]}, 0)\\
        & -\lambda \cdot \frac{1}{|N| \cdot m} \sum_{n \in N} f(c_n, c_u)
    \end{split}
    \end{align}
    \label{thm:3}
    \end{footnotesize}
\end{theorem}

The probability term $P(s_{n, [K]} = q_{n, [i]})$ for $n \in N$ is defined over all random subsets $S$, where $s_{n, [K]}$ could be seen as a random variable. This probability term could be solved using dynamic programming in $O(K \cdot |B| \cdot |N|)$ time complexity by the following recursion equations:
\begin{footnotesize}
\begin{align}
\label{eqn:dp}
\begin{split}
    \left\{
    \begin{aligned}
    P(s_{n, [j]}  &\geq  q_{n, [i]}) = (1 - x_{[i]}) \cdot P(s_{n, [j]} \geq q_{n, [i-1]}) \\
    &\;+ x_{[i]} \cdot  P(s_{n, [j-1]} \geq q_{n, [i-1]}) \; for \; [i] \in B_u\\
    P(s_{n, [j]} &\geq q_{n, [i]}) = P(s_{n, [j-1]} \geq q_{n, [i-1]}) \; for \; [i] \in B_s
    \end{aligned}
    \right.\\
    P(s_{n, [j]} \! = \! q_{n, [i]}) \!= \!P(s_{n, [j]} \geq q_{n, [i]}) - P(s_{n, [j]} \geq q_{n, [i-1]})
\end{split}
\end{align}
\end{footnotesize}

where $j$ runs for $1 \cdots K$ and $i$ runs for $1 \cdots |B|$. \footnote{Details are shown in Appendix 2.2 and 2.3.}

Algorithm \ref{alg:D} shows the complete process of the Continuous Double Greedy Algorithm. The algorithm first uses a gradient-based method to optimize the surrogate multilinear extension of the submodular target function and returns a sub-optimal continuous vector $x$, which represents for the probability each element is selected. Then, certain rounding technique such as Pipage Rounding \cite{calinescu2011maximizing, vondrak2013symmetry} is used to transform the resulting fractional solution into an integral solution. \footnote{See Appendix 2.4.}

\begin{algorithm}
    \caption{Continuous Double Greedy Algorithm ($F$, $m$)}
    \begin{algorithmic}[1]
        \STATE {Initialize: $x^0 \leftarrow \emptyset$, $y^0 \leftarrow B_u$} 
        \FOR{time step $t \in [1, T]$}
            \FOR {every $u \in B_u$}
                \STATE {Let $a_u \! \leftarrow \! \frac{\partial F(x^{t-1})}{\partial x_u}$, $b_u \! \leftarrow \! \frac{\partial F(y^{t-1})}{\partial y_u}$ by Eq. \ref{eqn:derivative}, \ref{eqn:dp}.}
                \STATE {Let $a_u^{\prime}(l) \!\leftarrow\! max(a_u \!- l, 0)$,$b_u^{\prime}(l) \!\leftarrow\! max(b_u \!+ l, 0)$}
                \STATE {Let $\frac{dx_u}{dt}(l, t\!-\!1)\!\leftarrow\!\frac{a_u^{\prime}}{a_u^{\prime} + b_u^{\prime}}$, $\frac{dy_u}{dt}(l, t-1)\!\leftarrow\!-\frac{b_u^{\prime}}{a_u^{\prime} + b_u^{\prime}}$.}
            \ENDFOR
            \STATE {Find $l^{*}$ satisfying $\sum_{u \in B_u} \frac{dx_u}{dt}(l^{*}, t-1)=m$.} 
            \STATE {Do a step of Gradient Ascent for $x$ and Gradient Descent for $y$: $x_u^t = x_u^{t-1} + \frac{1}{T} \cdot \frac{dx_u}{dt}(l^{*}, t-1)$, $y_u^t = y_u^{t-1} - \frac{1}{T} \cdot \frac{dy_u}{dt}(l^{*}, t-1)$.}
        \ENDFOR
        \STATE Process certain rounding technique using $x^T$ to get $U$.
        \RETURN {$U$}
    \end{algorithmic}
    \label{alg:D}
\end{algorithm}

A similar optimization bound analysis of Algorithm \ref{alg:C} and Algorithm \ref{alg:D} is given in Theorem \ref{thm:4}.

\begin{theorem}
    For $B_s = \emptyset$ and $\lambda > 0$, using a combination of Algorithm \ref{alg:C} and \ref{alg:D} to solve for optimization problem \ref{eq:opt} with $\lambda > 0$, $h$ and $Q$ are defined same as Theorem \ref{thm:2}, we have 
    \begin{footnotesize}
    \[
        \begin{split}
        \mathbb{E}(h(U)) \geq &\;max\;(\frac{1-m/er}{e} \cdot h(OPT) + C_1 \cdot Q, \\
        &(1 + \frac{r}{2\sqrt{(r-m)m}})^{-1} \cdot h(OPT) + C_2 \cdot Q)
        \end{split}
    \]
    \end{footnotesize}
    For $0 < \lambda < \frac{1}{e-1}$, we have $C_1 = \frac{1}{e} + (1 - \frac{1}{e}) \frac{m}{r} - (1 - \frac{1}{e}) \cdot \lambda > 0$ and $C_2 = \frac{(1 - \lambda) r}{2 \sqrt{(r-m)m} + r} - \epsilon \geq \frac{1}{2}(1 - \lambda) > 0$,where $r=|B_u|$. The first term is the lower bound for Algorithm \ref{alg:C} and the second term for Algorithm \ref{alg:D}.
    \label{thm:4}
\end{theorem}

Theorem \ref{thm:4} indicates that if neglecting the term with $Q$, when $m < 0.08r$ or $m > 0.92r$, we should use Algorithm \ref{alg:C} and otherwise Algorithm \ref{alg:D} by comparing two bounds.

\begin{table*}[htbp]
    \caption{\label{tab:algorithm} Conclusion of Applicability of Different Algorithms}
    \centering
    \scalebox{0.8}{
    \begin{tabular}{cccc}
    \toprule
    \textbf{Parameter} & \textbf{Algorithm} & \textbf{Applicability} & \textbf{Complexity} \\
    \midrule
    $\lambda=0$ & Greedy on Novel Class & $m > \gamma \cdot K \cdot |N|$, with $\gamma$ slightly larger than 1 & $O(|B| \cdot log|B| \cdot |N|)$ \\
    $\lambda=0$ & Greedy on Target Function & $m < \gamma \cdot K \cdot |N|$, with $\gamma$ slightly larger than 1 & $O(m \cdot (|B| + |N| \cdot logK))$ \\
    $\lambda>0$ & Random Greedy & $m < 0.08 \cdot |B_u|$ or $m > 0.92 \cdot |B_u|$ & $O(m \cdot (|B| \cdot log m + |N| \cdot logK))$ \\
    $\lambda>0$ & Continuous Double Greedy & $0.08 \cdot |B_u| < m < 0.92 \cdot |B_u|$ & $O(T \cdot K \cdot |B|^2 \cdot |N|)$ \\
    \bottomrule
    \end{tabular}}
\end{table*}

As a conclusion of this section, we list the applicability of different algorithms for this specific problem in Table \ref{tab:algorithm}.

\section{Experiments}
\subsection{Experimental Settings}
Basically, we design three different settings to show the superiority of our proposed algorithm: 

\noindent \textbf{Pre-trained Selection} A pre-trained model is given, and the base classes selection could be conducted with the help of the pre-trained model. Generally we could use the pre-trained model to extract image representations. The setting also supposes that we know about the novel support set. In this paper, we evaluate the generalized performance only via the base model trained on the selected base classes, while in practice we could also use these selected base classes to further fine-tune the given pre-trained model.

\noindent \textbf{Cold Start Selection} No pre-trained model is given, hence the base classes selection is conducted in an incremental manner. For each turn, the selection of the incremental base classes is based on the trained base model from the previous turn. The novel support set is also given. Note that the setting is somewhat like a curriculum learning \cite{Bengio2009Curriculum}.

\noindent \textbf{General Selection} The novel support set is not known beforehand (\textit{i.e.} Select a general base dataset that performs well on \textbf{any} composition of novel classes). In this paper for simplicity, we also suppose a pre-trained model is given as in the Pre-trained Selection setting.

In our experiments, we use two datasets for validating general classification: ImageNet and Caltech256, and one for fine-grained classification: CUB-200-2011. For ImageNet, we use the other 500 classes in addition to those used in the preliminary experiment in Section 3, which are further split into 400 candidate base classes and 100 novel classes. For all three tasks, the base dataset is selected from these 400 candidate base classes, and further evaluate the generalization performance on the 100 novel ImageNet classes, Caltech256 and CUB-200-2011.

For all experiments, we train a standard ResNet-18 \cite{he2016deep} backbone as the base model on the selected base classes. For few-shot learning task on novel classes, we use two different heads: one is the cosine similarity on the representation space (512-dimensional features after conv5\_x layer), which is a simplified version of Matching Network \cite{vinyals2016matching} without meta training step, representing the branch of metric-based approaches in few-shot learning. The other is the softmax regression on the representation space, which is a simple method from the branch of learning-based approaches. \footnote{The result of softmax regression head is shown in 4.} We use different heads to show our proposed selection method is model-agnostic.

As for the details of the experiment, we use an active learning manner as mentioned in Section 1. Each candidate base class only contains 50 images before selected. We utilize these images to calculate class representation. When a base class is selected, the number of training images for this class could be expanded to a relatively abundant number (For this experiment all training images of this class in ImageNet are used, which locates at the interval from about 800 to 1,300). We allow for a slight difference in the number of images per class to simulate a practical scenario. For a p-way k-shot setting, we randomly select p novel classes and then choose k samples per novel class as the novel support set; another 100, 50, 40 samples disjoint with the support set per novel class as the novel testing set for ImageNet, Caltech and CUB-200-2011. The flow of the experiment is to run selection algorithms, expand the selected classes, train a base model on the expanded base dataset and evaluate performance on testing set. The process is repeated for 10 times with different randomization, and we report the average Top-1 accuracy for each experiment setting. For settings containing pre-trained model, in this paper we use ResNet-18 trained on full training images from randomly selected 100 classes extracted from the candidate base classes in Section 3, which is disjoint with the base and novel classes used in this section. We also emphasize that when comparing with different methods within the same setting, the same novel support set and novel testing set are used for each turn of the experiment for a fair comparison.

We consider three baselines in our experiments: the first is the Random Selection, which draws the base classes uniformly, which is a rather simple baseline but common in the real scenario, the second is using the Domain Similarity metric which is generally used in \cite{remus2012domain, plank2011effective, ruder2017learning}. The idea is to maximize a pre-defined domain similarity between representation for each selected element in the source domain and the representation for the target domain. The method is first proposed for sample selection, and in this paper we extend to the class selection by viewing the centroid of features for a class as a sample and viewing the centroid of the novel support set as representation for the target domain. The baseline will be used in Pre-trained Selection and Cold Start Selection. The third is the K-medoids algorithm \cite{HaeA}, which is a clustering algorithm as a baseline of the General Selection setting. For all baselines and our algorithm, cosine similarity on representation space is used for calculating the similarity of two representations.

\begin{table*}[htbp]
    \caption{\label{tab:exp1imagenet} ImageNet: Pre-trained Selection, 100-way novel classes}
    \centering
    \scalebox{0.8}{\begin{tabular}{ccccc}
    \toprule
    \textbf{Algorithm} & \textbf{m=100, 5-shot} & \textbf{m=100, 20-shot} & \textbf{m=20, 5-shot} & \textbf{m=20, 20-shot}\\
    \midrule
    Random & $39.39\% \pm 0.82\%$ & $49.47\% \pm 0.67\%$ & $23.89\% \pm 0.56\%$ & $33.06\% \pm 0.47\%$ \\
    DomSim & $38.00\% \pm 0.36\%$ & $48.80\% \pm 0.79\%$ & $23.15\% \pm 0.43\%$ & $31.81\% \pm 0.58\%$ \\
    Alg. 1, \text{$K=1, \lambda=0$} & $\mathbf{43.42\% \pm 0.78\%}$ & $\mathbf{53.79\% \pm 0.37\%}$ & $25.71\% \pm 0.43\%$ & $34.67\% \pm 0.36\%$ \\
    Alg. 2, \text{$K=1, \lambda=0$} & $43.20\% \pm 0.76\%$ & $53.61\% \pm 0.27\%$ & $\mathbf{26.13\% \pm 0.44\%}$ & $\mathbf{34.97\% \pm 0.45\%}$ \\
    Alg. 2, \text{$K=3, \lambda=0$} & $42.89\% \pm 0.43\%$ & $53.13\% \pm 0.27\%$ & $25.10\% \pm 0.48\%$ & $34.52\% \pm 0.51\%$ \\
    
    \bottomrule
    \end{tabular}}
\end{table*}

\begin{table}[htbp]
    \caption{\label{tab:exp1caltech} Caltech256: Pre-trained Selection, 100-way}
    \centering
    \scalebox{0.8}{\begin{tabular}{ccc}
    \toprule
    \textbf{Algorithm} & \textbf{m=100, 5-shot} & \textbf{m=100, 20-shot}\\
    \midrule
    Random & $45.31\% \pm 1.32\%$ & $54.97\% \pm 1.23\%$\\
    DomSim & $49.55\% \pm 1.28\%$ & $58.84\% \pm 1.01\%$\\
    Alg. 1, \text{$K=1, \lambda=0$} & $\mathbf{55.41\% \pm 1.25\%}$ & $\mathbf{64.46\% \pm 0.99\%}$\\
    Alg. 2, \text{$K=3, \lambda=0$} & $54.94\% \pm 1.14\%$ & $63.58\% \pm 0.98\%$\\
    
    \bottomrule
    \end{tabular}}
\end{table}

\begin{table}[htbp]
    \caption{\label{tab:exp1cub} CUB-200-2011: Pre-trained Selection, 100-way}
    \centering
    \scalebox{0.8}{\begin{tabular}{ccc}
    \toprule
    \textbf{Algorithm} & \textbf{m=100, 5-shot} & \textbf{m=100, 20-shot}\\
    \midrule
    Random & $18.46\% \pm 1.19\%$ & $26.14\% \pm 1.44\%$\\
    DomSim & $28.11\% \pm 0.44\%$ & $38.26\% \pm 0.45\%$\\
    Alg. 1, \text{$K=1, \lambda=0$} & $\mathbf{29.65\% \pm 0.82\%}$ & $\mathbf{39.77\% \pm 0.41\%}$\\
    Alg. 2, \text{$K=3, \lambda=0$} & $28.04\% \pm 1.82\%$ & $37.22\% \pm 0.69\%$\\
    
    \bottomrule
    \end{tabular}}
\end{table}

\subsection{Results}
\subsubsection{Pre-trained Selection}
Table \ref{tab:exp1imagenet}, \ref{tab:exp1caltech}, \ref{tab:exp1cub} show the results of the Pre-trained Selection. When setting $K=1$, the algorithm reaches the best performance in all cases. For the ImageNet dataset in Table \ref{tab:exp1imagenet}, we show that Algorithm 1 and Algorithm 2 are fit for different cases, depending on the number of selected classes, as Table \ref{tab:algorithm} describes. For $m=100$ and $m=20$ case, our algorithm obtains a superior accuracy of about 4\% and 2\% separately compared with random selection, which is a relatively huge promotion in few-shot image classification. Besides, the promotion is rather stable concerning the shot number. The Domain Similarity algorithm performs worse because of the cluster effect, where the selected base classes are concentrated around the centroid of the target domain, in contrast with the idea of enhancing diversity we show in Section 3. For Caltech256 as novel classes in Table \ref{tab:exp1caltech}, a transfer distribution on dataset is introduced. It shows that in such case, the improved margin compared to random selection is much larger, reaching about 10\% when $m=100$. This is because our algorithm enjoys the double advantages of transfer effect and class selection effect; the former also promotes the Domain Similarity algorithm. For the CUB-200-2011 dataset in Table \ref{tab:exp1cub}, we further show that our algorithm improves the margin much more significantly in a fine-grained manner, reaching about 11.2\% for 5-shot setting and 13.6\% for 20-shot setting.

\begin{table}[htbp]
    \caption{\label{tab:exp2imagenet} ImageNet: Cold Start Selection, 100-way}
    \centering
    \scalebox{0.8}{\begin{tabular}{ccccc}
    \toprule
    \textbf{Algorithm} & \textbf{Mechanism} & \textbf{Top-1 Accuracy}\\
    \midrule
    Random & - & $39.39\% \pm 0.82\%$\\
    DomSim & 6-12-25-50-100 & $39.30\% \pm 0.40\%$\\
    Alg. 1, \text{$K=1, \lambda=0$} & 6-12-25-50-100 & $40.96\% \pm 0.50\%$\\
    Alg. 2, \text{$K=1, \lambda=0$} & 6-12-25-50-100 & $41.75\% \pm 0.59\%$\\
    Alg. 2, \text{$K=3, \lambda=0$} & 6-12-25-50-100 & $\mathbf{42.17\% \pm 0.67\%}$\\
    Alg. 2, \text{$K=5, \lambda=0$} & 6-12-25-50-100& $41.33\% \pm 0.36\%$\\
    Alg. 2, \text{$K=3, \lambda=0$} & 10-20-40-80-100 & $41.61\% \pm 0.76\%$\\
    Alg. 2, \text{$K=3, \lambda=0$} & 50-100 & $40.88\% \pm 0.66\%$\\
    \hline
    Pre-trained (Upperbound) & [100]-100 & $42.89\% \pm 0.43\%$\\
    
    \bottomrule
    \end{tabular}}
\end{table}

\vspace{-0.3cm}
\subsubsection{Cold Start Selection}
The Cold Start Selection is more difficult than the Pre-trained Selection in that there is no pre-trained model at the early stage, leading to an unknown or imprecise image representation space. Hence the representation space needs to be learned incrementally. For each turn, the selection of the incremental base classes is based on the trained base model from the previous turn. Noticing that in this incremental learning manner both the complexity and the effectiveness of selection should be considered. To limit the complexity we increasingly select the same number of classes in each turn as the total number of selected base classes in the previous turn (\textit{i.e.} doubling the number of selected classes in each turn). This double-increasing mechanism could guarantee a linear time complexity of $m$ in training the base model. For example, in Table \ref{tab:exp2imagenet} a 6-12-25-50-100 mechanism represents for selecting 6 classes randomly in Turn 1, and continue selecting another 6 classes based on the model trained by classes from Turn 1 to form a selection of 12 classes in Turn 2 and so on. As the representation space is not so stable as the Pre-trained Selection, a larger K with $K=3$, $\lambda=0$ is much better. Table \ref{tab:exp2imagenet} shows the result of the algorithms. Our proposed method exhibits a 2.8\% promotion compared to random selection. We also highlight that the upper bound of the algorithm is limited by the Pre-trained selection (with a pre-trained model on 100 classes with $K=3$), which is 42.89\%. By using the double-increasing mechanism, the performance is just slightly lower than this upper bound in linear time complexity.

We also show some ablation studies by changing the selection of $K$ and the selection mechanism. As for the selection mechanism, comparing 6-12-25-50-100 and 50-100, we draw a conclusion that the incremental learning of the representation space is much more effective, and compared to 10-20-40-80-100 it shows that the selection in the early stage of Cold Start Selection is more important than the later stage.

\vspace{-0.3cm}
\subsubsection{General Selection}
General Selection is the most difficult setting in this paper, as we do not know the novel classes previously. The goal is to select a base dataset that could perform well on any composition of novel classes. In dealing with this problem, we make a slight change to our optimization framework that \textbf{we take all candidate base classes as the novel classes}. The implicit assumption is that the candidate base classes represent for a subsample of the global world categories. In this setting, we should choose a much larger $K$ and $\lambda$ for this setting, especially for fine-grained classification, to enhance representativeness and diversity for each selected class.

Results of ImageNet and Caltech256 (Table \ref{tab:exp3imagenet}, \ref{tab:exp3caltech}) show that our algorithms perform much better when the number of selected classes is larger. Specifically, in $m=100$ case we promote 0.9\% and 4.5\% in two datasets separately compared with random selection, however in $m=20$ case the promotion is not so obvious, only 0.3\% and 0.9\%, which shows that a larger base dataset may contain more general image information. As for the result of CUB-200-2011 (Table \ref{tab:exp3cub}), our proposed algorithm performs much better due to the effect of diversity, reaching an increase of 6.4\% in $m=100$ case. Besides, the result also shows that the performance reaches the best with a positive $\lambda$ in fine-grained classification, illustrating the necessity of diversity (According to Table \ref{tab:algorithm}, we choose Algorithm 3 for $m=20$ and Algorithm 4 for $m=100$). The results also show that the baseline K-Medoids is rather unstable in different cases. It may reach the state-of-the-art in some cases but may perform even worse than random in other cases.

\begin{table}[htbp]
    \caption{\label{tab:exp3imagenet} ImageNet: General Selection, 100-way}
    \centering
    \scalebox{0.8}{\begin{tabular}{ccc}
    \toprule
    \textbf{Algorithm} & \textbf{m=20, 20-shot} & \textbf{m=100, 20-shot}\\
    \midrule
    Random & $33.06\% \pm 0.47\%$ & $49.47\% \pm 0.67\%$\\
    K-Medoids & $\mathbf{33.50\% \pm 0.28\%}$ & $49.17\% \pm 0.38\%$\\
    Alg. 2, \text{$K=3, \lambda=0$} & $33.38\% \pm 0.25\%$ & $50.00\% \pm 0.38\%$\\
    Alg. 2, \text{$K=5, \lambda=0$} & $33.32\% \pm 0.30\%$ & $\mathbf{50.35\% \pm 0.29\%}$\\
    Alg. 2, \text{$K=10, \lambda=0$} & $33.01\% \pm 0.38\%$ & $50.21\% \pm 0.26\%$\\
    Alg. 3/4, \text{$K=5, \lambda=0.2$} & $32.82\% \pm 0.35\%$ & $49.19\% \pm 0.34\%$ \\
    
    \bottomrule
    \end{tabular}}
\end{table}

\vspace{-0.1cm}
\begin{table}[htbp]
    \caption{\label{tab:exp3caltech} Caltech256: General Selection, 100-way}
    \centering
    \scalebox{0.8}{\begin{tabular}{ccc}
    \toprule
    \textbf{Algorithm} & \textbf{m=20, 20-shot} & \textbf{m=100, 20-shot}\\
    \midrule
    Random & $40.26\% \pm 0.90\%$ & $54.97\% \pm 1.23\%$\\
    K-Medoids & $40.16\% \pm 0.83\%$ & $59.27\% \pm 1.01\%$\\
    Alg. 2, \text{$K=3, \lambda=0$} & $40.72\% \pm 0.92\%$ & $59.23\% \pm 0.94\%$\\
    Alg. 2, \text{$K=5, \lambda=0$} & $40.98\% \pm 0.84\%$ & $58.68\% \pm 0.94\%$\\
    Alg. 2, \text{$K=10, \lambda=0$} & $\mathbf{41.18\% \pm 0.88\%}$ & $\mathbf{59.52\% \pm 0.91\%}$\\
    Alg. 3/4, \text{$K=5, \lambda=0.2$} & $40.31\% \pm 1.24\%$ & $57.79\% \pm 0.94\%$ \\
    
    \bottomrule
    \end{tabular}}
\end{table}

\vspace{-0.1cm}
\begin{table}[htbp]
    \caption{\label{tab:exp3cub} CUB-200-2011: General Selection, 100-way}
    \centering
    \scalebox{0.8}{\begin{tabular}{ccc}
    \toprule
    \textbf{Algorithm} & \textbf{m=20, 20-shot} & \textbf{m=100, 20-shot}\\
    \midrule
    Random & $15.25\% \pm 0.91\%$ & $26.14\% \pm 1.44\%$\\
    K-Medoids & $14.96\% \pm 0.47\%$ & $24.38\% \pm 0.59\%$\\
    Alg. 2, \text{$K=3, \lambda=0$} & $14.74\% \pm 0.48\%$ & $27.08\% \pm 0.52\%$\\
    Alg. 2, \text{$K=5, \lambda=0$} & $16.06\% \pm 0.59\%$ & $28.33\% \pm 0.57\%$\\
    Alg. 2, \text{$K=10, \lambda=0$} & $16.21\% \pm 0.33\%$ & $27.63\% \pm 0.66\%$\\
    Alg. 3/4, \text{$K=5, \lambda=0.2$} & $16.61\% \pm 0.36\%$ & $\mathbf{32.50\% \pm 0.58\%}$ \\
    Alg. 3/4, \text{$K=5, \lambda=0.5$} & $\mathbf{17.09\% \pm 0.33\%}$ & $31.01\% \pm 0.58\%$ \\
    
    \bottomrule
    \end{tabular}}
\end{table}

\section{Conclusions}

This paper focuses on how to construct a high-quality base dataset with limited number of classes from a wide broad of candidates. We propose the Similarity Ratio as a proxy of the performance of few-shot learning and further formulate the base class selection problem as an optimization process over Similarity Ratio. Further experiments in different scenarios show that the proposed algorithm is superior to random selection and some typical baselines in selecting a better base dataset, which shows that, besides advanced few-shot algorithms, a reasonable selection of base dataset is also highly desired in few-shot learning.

\vspace{-0.3em}
\section{Acknowledgement}
This work was supported in part by National Key R\&D Program of China (No. 2018AAA0102004), National Natural Science Foundation of China (No. U1936219, No. 61772304,  No. U1611461), Beijing Academy of Artificial Intelligence (BAAI).